\documentclass{article}
\usepackage{iclr2027_conference,times}
\usepackage{amsmath,amssymb}
\usepackage{enumitem}
\usepackage{booktabs}
\usepackage{graphicx}
\usepackage{wrapfig}
\usepackage{array}
\usepackage{multirow}
\usepackage{xcolor}
\usepackage{tikz}
\usetikzlibrary{arrows.meta,positioning,fit,backgrounds,calc,decorations.pathreplacing,shadows}
\usepackage[colorlinks=true,linkcolor=black,citecolor=blue,urlcolor=blue]{hyperref}
\usepackage{url}
\usepackage[capitalise,noabbrev]{cleveref}   %

\newcommand{\FarmVisionOnlyOodNclips}{73}

\newcommand{\FarmVisionOnlyIndomNclips}{200}

\newcommand{\FarmSubMethodSubNclips}{3{,}000}

\newcommand{\FarmSubMethodSidNclips}{2{,}728}

\newcommand{\FarmSubMethodSoodNclips}{272}

\newcommand{\DconMainOodLpipsRel}{5.1\%}

\newcommand{\DconMainOodLpipsLocalRel}{7.4\%}

\newcommand{\DconSigmaWidthOodLpips}{\ensuremath{+0.0005}}

\newcommand{\DconSubMainSidLpipsRel}{2.0\%}

\newcommand{\DconSubMainSoodLpipsRel}{2.2\%}

\newcommand{\DconPwTWVsMethodSubLpipsRel}{14.7\%}

\newcommand{\DconPwHGVsMethodSubLpipsRel}{1.8\%}

\newcommand{\RunPairsOodNeg}{7}
\newcommand{\RunPairsOodN}{9}

\newcommand{\RunPairsIndomNeg}{7}
\newcommand{\RunPairsIndomN}{9}

\newcommand{\HandMethodPVsControlPFullFlowMagGapB}{23\%}
\newcommand{\HandMethodPVsControlPFullFlowMagGapA}{9\%}

\newcommand{\BlockRatioOod}{4.1}

\newif\ifHaveBitThree\HaveBitThreefalse
\newif\ifHaveWThreeSeven\HaveWThreeSeventrue
\newif\ifHaveOracleSwap\HaveOracleSwaptrue
\newif\ifHaveHandMain\HaveHandMaintrue
\newif\ifHaveHandBit\HaveHandBittrue

\newcommand{\CorpusEpisodes}{1{,}926}
\newcommand{\CorpusClips}{43{,}604}

\newcommand{\SplitTrainClips}{38{,}600}

\newcommand{\SharpControlRatio}{0.146}

\newcommand{\SharpMethodRatio}{0.171}

\newcommand{\SharpMethodN}{3{,}000}
\newcommand{\SharpGateZeroRatio}{0.167}

\newcommand{\SharpMethodVsControlDelta}{\ensuremath{+0.025}}
\newcommand{\SharpMethodVsControlCi}{[\ensuremath{+0.024},\,\ensuremath{+0.026}]}

\newcommand{\SanityN}{150}

\newcommand{\SanityDtwmLpips}{0.412}
\newcommand{\SanityDtwmPsnr}{14.94}
\newcommand{\SanityDtwmSsim}{0.535}

\newcommand{\SanityBlurThreeLpips}{0.625}
\newcommand{\SanityBlurThreePsnr}{15.30}
\newcommand{\SanityBlurThreeSsim}{0.571}

\newcommand{\SanityBlurThreePsnrWin}{98\%}
\newcommand{\SanityBlurThreeSsimWin}{100\%}

\newcommand{\ProbeN}{150}
\newcommand{\ProbeRawGap}{\ensuremath{-0.30}}

\newcommand{\ProbeGainShare}{61\%}
\newcommand{\ProbeBlurShare}{9\%}
\newcommand{\PhotoN}{150}

\newcommand{\PhotoBaselineStd}{\ensuremath{-7.8}}
\newcommand{\PhotoBaselineSat}{\ensuremath{-16.4}}

\newcommand{\PhotoDtwmStd}{\ensuremath{-3.9}}
\newcommand{\PhotoDtwmSat}{\ensuremath{-11.0}}

\newcommand{\ForceNClips}{12{,}000}
\newcommand{\ForceOneMerged}{5\%}
\newcommand{\ForceOneFlip}{17\%}
\newcommand{\ForceOneChangeTwenty}{46\%}
\newcommand{\ForceOneChangeFifty}{22\%}
\newcommand{\ForceOneCopMove}{48\%}

\newcommand{\ForceOneReleaseDec}{16\%}
\newcommand{\ForceOneReleaseSteady}{7\%}
\newcommand{\ForceOneReleaseInc}{12\%}
\newcommand{\ForceOneRatioMin}{1.2}
\newcommand{\ForceOneRatioMax}{4.7}
\newcommand{\ForceOneAucBase}{0.723}
\newcommand{\ForceOneAucFull}{0.735}
\newcommand{\ForceOneAucGain}{+0.012}
\newcommand{\ForceOneAucGainCi}{[+0.007, +0.018]}

\newcommand{\ForceTwoAucBase}{0.627}
\newcommand{\ForceTwoAucFull}{0.649}

\newcommand{\ForceDecVsSteady}{2.5}
\newcommand{\ForceIncVsSteady}{1.8}

\newcommand{\method}{DTWM}
\newcommand{\corpus}{EgoTouch}

\title{Dexterous Tactile World Model}

\author{Ziyao Zeng$^{1}$\quad Xiatao Sun$^{1}$\quad Hao Wang$^{2}$\quad Yueyang Pan$^{3}$\quad Zhengxiang Yu$^{4}$\\
\textbf{Fengyu Yang$^{1}$\quad Tianyu Liu$^{1}$\quad Zhiwen Fan$^{2}$\quad Daniel Rakita$^{1}$}\\[4pt]
\normalfont $^{1}$Yale University\quad $^{2}$Texas A\&M University\\
\normalfont $^{3}$University of California, Los Angeles\quad $^{4}$University of Washington}

\iclrfinalcopy

\begin{document}
\maketitle
\lhead{Preprint}
{\renewcommand{\thefootnote}{}\footnotetext{Project page: \url{https://adonis-galaxy.github.io/dtwm-project-page/}}}

\begin{abstract}
World models for manipulation are typically trained from video, yet the events that determine how manipulation unfolds, such as making and releasing contact, are difficult to observe visually and are often easier to sense through touch. We present the Dexterous Tactile World Model (\method{}), a video world model for future-frame prediction of egocentric manipulation from both observed video and tactile signals from a glove worn on each hand. We condition a pretrained video diffusion transformer on each hand's tactile signal through a zero-initialized residual at the corresponding hand location in the video tokens, while a causal mask prevents predicted frames from accessing future information. Compared with a vision-only model matched in architecture, parameters, and training, \method{} reduces the underestimation of hand motion from \HandMethodPVsControlPFullFlowMagGapB{} to \HandMethodPVsControlPFullFlowMagGapA{}, while reducing the perceptual error in the hand region by \DconMainOodLpipsLocalRel{} across three training runs per model. The benefit also increases over the prediction horizon, with the improvement in the later predicted chunks being about \BlockRatioOod{}x larger than in the first. \method{} also outperforms other visual-tactile world models under the same setting, and training with touch improves future-frame prediction even when no touch is available at inference. Ablations show that the model benefits from both the magnitude and spatial location of force: replacing the tactile signal with binary contact states, either per hand or per location, increases prediction error. The observed course of the force indicates whether the interaction will persist or change.

\end{abstract}

\section{Introduction}
\label{sec:intro}
\suppressfloats[t]

\begin{figure}[t]
\centering
\includegraphics[width=\textwidth]{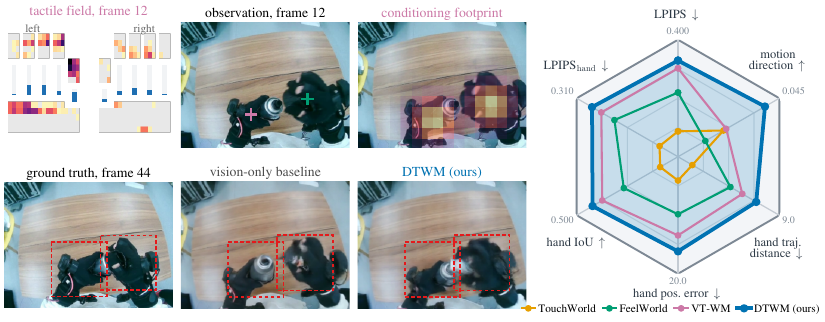}
\caption{\textbf{\method{} shows better future-frame prediction of human manipulation with whole-hand
tactile sensing.} Left: a held-out clip in which both hands open a spray can; dashed boxes are the
ground-truth hand regions. Right: we achieve better visual quality of the entire frame and the hand crop,
and more accurate hand location and motion, than existing visual-tactile world models.}
\label{fig:teaser}
\end{figure}

A world model predicts how the physical world will evolve from past observations and other conditions, typically as future frames. Such models are useful
for physical AI in three ways. They can \emph{synthesize data}: one recording can yield many possible futures, since real-world manipulation data is costly to scale \citep{yang2024unisim,jang2025dreamgen,agarwal2025cosmos,liu2026openlongtail}. They can serve as
\emph{initialization}: a model that predicts the future well implicitly models how objects
interact, which makes it a strong initialization for policies and other models
of physical interaction \citep{wu2024gr1,hu2025vpp}. They can also serve as an \emph{environment}: a robot
policy \citep{kim2024openvla,black2025pi0,sun2026decoupling,sun2026foveated} can be trained or evaluated in the world model instead of the real world
\citep{ha2018recurrent,hafner2020dream,bruce2024genie,quevedo2025worldgym}.

All three uses require accurate predictions where the outcome is decided: at contact. In manipulation, whether a grasp holds, an object slips, or a finger reaches a surface determines what happens next. Yet world models are trained mainly from video, where contact is hard to observe. The hand often occludes the contact region, and touching an object can look nearly identical to pressing firmly against it. The model must therefore infer contact from ambiguous visual evidence, and its uncertainty propagates to the predictions. The failure is visible in the generated video: predicted hands may lose their fingers, merge with the objects they hold, or move less than they should (\Cref{fig:teaser,fig:id}).

Touch provides a direct signal of these events. A pressure-sensing glove records where and how firmly each hand makes contact, including contacts that are hidden from the camera. Recent datasets pair egocentric videos of everyday manipulation with such tactile measurements from both hands \citep{zhou2026touchanything}. We therefore condition a world model on dexterous touch: given the observed video frames and the tactile signals from both hands, it predicts how the manipulation unfolds next.

Several recent world models incorporate touch into video prediction
\citep{higuera2026visuotactile,ma2026feelworld,huang2026vitacworld,zhou2026touchworld}, but differ from ours
in data, setting, and model design. Three are trained on robot trajectories from
fingertip or gripper-mounted sensors \citep{higuera2026visuotactile,ma2026feelworld,huang2026vitacworld},
whereas TouchWorld \citep{zhou2026touchworld} uses the same human corpus as ours before adapting to a pair of
robot hands. All four evaluate touch in robot settings, where the world model is conditioned on the robot's
actions or on a subtask and serves planning or a policy. We instead predict egocentric human manipulation
from past observations alone, so the model must also infer how the hands themselves will move. Finally,
they represent touch as global tactile features, additional tokens or a separate tactile image, none
aligned with the image region where contact occurs. We call this missing alignment the
\emph{indexing mismatch}: a glove reading is indexed by patches of skin, a video transformer by places in
the image, and the touched skin can be anywhere in the frame.

\method{} resolves the mismatch by placing the observed tactile readings at the observed hand location, calculated using rendered hand skeleton (\Cref{fig:pipeline}). We embed each hand's reading, spread it over the video tokens
around that location with a Gaussian footprint, and add it as a residual inside a pretrained video
diffusion transformer. The residual is computed by a zero-initialized projection, so the pretrained model
is unchanged at initialization and the pathway is learned rather than imposed. The reading is supplied
only for the observed frames, and a causal mask keeps every predicted frame from attending to later ones.
The model therefore predicts from past observations alone, with no future hand pose, action or reading, as
it would at deployment. The backbone, the autoencoder and the text encoder stay frozen, and the tactile
pathway adds only a few linear maps.

We evaluate \method{} by observing thirteen frames and predicting the next
thirty-six in one pass. The vision-only baseline is matched
in architecture, parameters and training, with its tactile input zeroed, so the two models differ only in whether the pathway carries the sensor's signal. The vision-only model underestimates how
much the hands move by \HandMethodPVsControlPFullFlowMagGapB{}, and \method{} by
\HandMethodPVsControlPFullFlowMagGapA{}. The perceptual error in the hand region falls by
\DconMainOodLpipsLocalRel{} across three training runs per model, against \DconMainOodLpipsRel{} over the
whole frame. The improvement also grows along the prediction and is about \BlockRatioOod{}x larger in the
later chunks than in the first. We also outperform three existing visual-tactile world models under the same setting, with a perceptual error lower by \DconPwHGVsMethodSubLpipsRel{} to
\DconPwTWVsMethodSubLpipsRel{}.

Ablations of the tactile input show that replacing the full signal with binary contact states, either at each location or for each hand, increases prediction error, with larger reductions in tactile information leading to larger errors. The model therefore uses not only whether a hand is in contact, but also where it is pressed and how firmly. We attribute this to the predictive setting: the observed tactile history captures how force evolves over time, rather than only its value at the end of the observation window. In the corpus, a hand whose force holds steady tends to keep its grasp, whereas a hand whose force decays or rises is \ForceIncVsSteady{} to \ForceDecVsSteady{} times as likely to release it within the next half second (\Cref{fig:motivation}), and this trend predicts a release beyond what the current force level does. Our contributions are the following:

\begin{itemize}[leftmargin=1.4em,itemsep=2pt,topsep=3pt]
\item \textbf{Dexterous tactile world model.} We propose \method{}, a world model for future-frame prediction of egocentric human manipulation. We fine-tune a pretrained video diffusion transformer on egocentric video and tactile signals from both hands, adding each hand's tactile signal to the video tokens at that hand's location through a Gaussian footprint.
\item \textbf{Improving future-frame prediction of manipulation.} Against a matched vision-only model, it
reduces the underestimation of hand motion from \HandMethodPVsControlPFullFlowMagGapB{} to
\HandMethodPVsControlPFullFlowMagGapA{} and the hand-region perceptual error by
\DconMainOodLpipsLocalRel{}. It also outperforms existing visual-tactile world models in the same setting.
\item \textbf{Dexterous tactile matters.} Conditioning on the location and magnitude of the tactile signal improves
prediction more than a binary contact state does. In the training data, how the force changes during the observation indicates whether the interaction will persist or change. Training with touch also improves prediction even when no touch is available at inference.
\end{itemize}

\section{Related Work}
\label{sec:related}

\paragraph{Dexterous tactile sensing.} Touch is sensed either at high resolution over a small pad or at
lower resolution over the whole hand. Optical sensors image the deformation of an elastomer and give dense
geometry at a fingertip \citep{yuan2017gelsight,lambeta2020digit}. Taxel arrays and tactile skins trade
that resolution for coverage of the palm and all fingers \citep{sundaram2019tactileglove}. Robot data comes
from teleoperating a gripper or a robot hand carrying such sensors, whereas wearable gloves record
people doing everyday tasks, and recent corpora pair egocentric video of bimanual activity with
synchronized pressure from both hands \citep{zhou2026touchanything,zhao2025egopressure}. These signals
serve grasp stability and slip detection \citep{calandra2018regrasp}, visuotactile
imitation \citep{huang2024vitac}, dexterous grasping without sight \citep{luo2026blind},
representation learning for control \citep{lee2019visiontouch,wang2020swingbot,fu2024touch}, and unified
tactile representations and generation across sensors and modalities \citep{yang2024unitouch,tu2026unitac}. These works use touch to recognize, grasp
or represent contact, not to predict how a manipulation unfolds; we instead feed whole-hand touch from
both hands to a video world model that predicts future frames.

\paragraph{Visual-tactile world models.} Video world models predict how a scene evolves from past frames,
first in pixel space \citep{oh2015actionconditional,finn2017visualforesight} and now mostly as latent video
diffusion transformers \citep{blattmann2023align,peebles2023scalable,wan2025video}, whose high perceptual
quality can still hide violations of physical laws \citep{chen2026crashtwin}. The fine-tuning of such models
can be controlled with an additional modality \citep{zhang2023controlnet,mou2024t2iadapter,zeng2025coffee}:
a global signal such as text or an action \citep{rombach2022latent,ye2023ipadapter,zeng2026iris},
a pixel-aligned image such as depth or pose, or extra channels concatenated at the input
\citep{blattmann2023stable}, usually added through zero-initialized layers \citep{zhang2024llamaadapter,hu2022lora}.
Several recent world models also condition video prediction on touch.
VT-WM \citep{higuera2026visuotactile} concatenates tokens of optical fingertip sensors with visual tokens in
an action-conditioned latent world model of a robot hand, and plans with it. FeelWorld
\citep{ma2026feelworld} lets visual tokens attend to one global token per fingertip sensor of a gripper,
through a contact-gated attention, and also plans with it. ViTacWorld
\citep{huang2026vitacworld} encodes optical tactile images from a gripper with the video autoencoder as an
additional view, and uses the model to generate data and to evaluate policies, and \citet{zhang2026contactworld} compare design
choices for such models. All of these are trained on robot data, with sensors on the fingertips of a
robot hand or on a gripper, whereas we train on egocentric videos of people wearing a
tactile glove on each hand.
TouchWorld \citep{zhou2026touchworld} uses the same corpus as ours: it adapts a pretrained video model to
this corpus and then to a pair of robot hands, where it supplies visual-tactile subgoals to a policy. Each of
the four is part of a robot system, used to plan actions, generate data and evaluate policies, or
supply subgoals to a policy, and all represent the tactile signal as extra tokens, global tactile features
or a separate tactile image, none of which is tied to the hand's location in the frame. \method{} is instead
a standalone world model of egocentric human manipulation, which predicts future frames from past
observations alone and adds each hand's whole-hand tactile reading to the video tokens at its location.

\section{Method}
\label{sec:method}

\begin{figure}[t]
\centering
\resizebox{\textwidth}{!}{\input{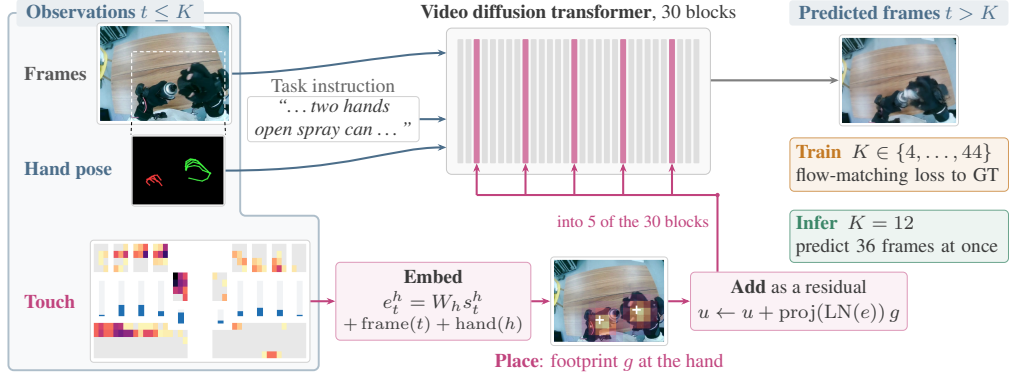}}
\caption{\textbf{Pipeline of \method{}.} During both training and inference, the model observes RGB frames, hand poses, and tactile signals from both hands for time steps $t \le K$. We sample one reading from both gloves ($255$ pressure taxels and $30$ finger-flexion channels in total) per video frame. In the touch panel, color shows pressure and blue bars show flexion. The task instruction is encoded by the text encoder, the observed frames are represented as clean latents, and the hand skeleton is encoded using a widened patch embedding. Each hand's tactile signal is embedded as $e$ and added to the video tokens at the hand's projected location using a normalized Gaussian footprint $g$:
$u \leftarrow u + \mathrm{proj}(\mathrm{LN}(e))\,g$.
The model then predicts the future RGB frames at time steps $t > K$. During training, $K$ is sampled from $\{4,\dots,44\}$, and the model is optimized with a flow-matching loss against the ground-truth future frames. At inference, the model observes 13 frames ($K=12$) and predicts all 36 future frames in a single pass.}
\label{fig:pipeline}
\end{figure}

\subsection{Problem setting}
\label{sec:method:setting}

We formulate manipulation prediction as block-causal chunked frame prediction with a pretrained
flow-matching \citep{lipman2023flow,liu2023rectifiedflow} video diffusion transformer
\citep{peebles2023scalable,wan2025video} with $30$ blocks and width $d=3072$. A causal video
autoencoder and a $(1,2,2)$ patchifier map a clip of $L=49$ frames at $384\times512$ to $T=13$ latent
frames on a $12\times16$ token grid, that is, $2496$ video tokens (\Cref{app:training}).

The prediction unit is a chunk of $C=4$ latent frames, or $16$ pixel frames. Attention is block-causal
\citep{yin2025causvid}: a token attends bidirectionally inside its own chunk and causally to earlier
chunks. Given the observation, pixel frames $0$ to $K$, that is latent frames $0$ to $m = K/4$, the
model denoises the remaining latents in one pass. Observed latents stay
clean with a per-token timestep of zero, and predicted latents share one sampled timestep
\citep{chen2024diffusionforcing}.

One pass therefore predicts three chunks jointly. The first predicted chunk attends only to observed
latents. Each later chunk also attends to the chunks before it, which are denoised in the same pass, so it
depends on the model's own predictions of them.
No chunk attends to anything that follows it. During training $K$ is drawn uniformly from
$\{4, 8, \dots, 44\}$. At deployment $K = 12$, the model predicts three chunks, or $36$ frames, and all of
them are evaluated (\Cref{sec:exp:setup}).

The conditioning contains only information available before the predicted interval (\Cref{fig:pipeline}).
The hand skeleton is encoded by a widened, zero-initialized patch embedding (\Cref{app:training}), and no
future hand pose or action is provided.

\subsection{Tactile conditioning at the hands}
\label{sec:method:injection}

The tactile signal is a pressure grid for each hand and frame, indexed somatotopically: each channel is a
fixed patch of skin. The transformer is indexed retinotopically, by its token grid. We
resolve the mismatch by giving each hand's signal the location of that hand in this grid.

\paragraph{Tactile representation.} For each pixel frame $t$ and hand $h$, the sensor provides a
hand-shaped $21\times21$ raster whose $217$ on-hand cells partly share raw glove channels. We remove the
duplicates. This leaves $138$ channels for the left hand and $147$ for the right. We standardize each
channel with fixed statistics and clip the result to $[-8, 8]$; some channels measure finger flexion
(\Cref{app:sensor}). The standardized state $s^h_t$ is embedded as
$e^h_t = W_h s^h_t + \mathrm{frame}(t) + \mathrm{hand}(h)$, using a linear map per hand and learned
frame and hand embeddings. Embeddings are averaged within each temporal group of the autoencoder, so
that $t$ subsequently indexes latent frames.

\Cref{sec:exp:input} replaces this signal with coarser representations, down to a single contact bit per
hand and frame, where a hand counts as being in contact when more than $2\%$ of its pressure taxels exceed
a normalized pressure of $0.05$ (\Cref{app:contact}).

\paragraph{Projecting the hands into the token grid.} The image location of each hand comes from the
rendered skeleton the model already receives, produced from per-frame 3D joints \citep{potamias2025wilor}
under a fixed pinhole projection. The pixel centroid of each hand's color gives normalized coordinates
$(y^h_t, x^h_t) \in [0,1]^2$ and a validity flag $v^h_t$, which is zero when the hand is not drawn. Both
are reduced to latent frames by a validity-weighted mean (\Cref{app:protocol}).

\begin{figure}[t]
\centering
\includegraphics[width=\textwidth]{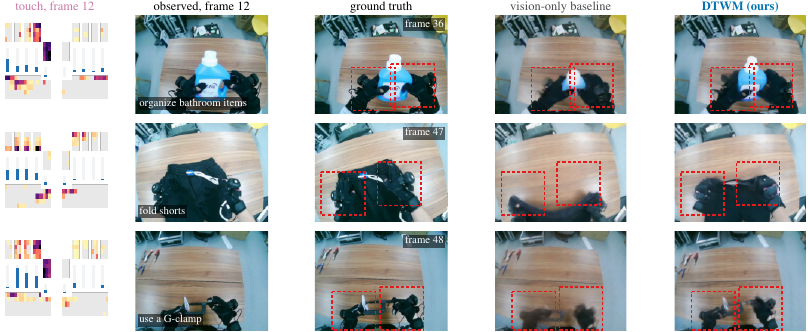}
\caption{\textbf{Comparison with the vision-only baseline.} The hands of the vision-only model lose their definition or
stop moving, while ours keep their shape and their motion. Warm colors: pressure, blue bars: finger
flexion. Dashed boxes: ground-truth hand regions.}
\label{fig:id}
\end{figure}

\paragraph{Footprint and residual.} At cell $p=(i,j)$ of the $H\times W$ token grid, each hand places an
isotropic Gaussian that is gated by validity and normalized to unit mass:
\begin{equation}
g^h_t(p) = \exp\!\Big(\!-\tfrac{(i - y^h_t H)^2 + (j - x^h_t W)^2}{2\sigma^2}\Big),
\qquad
\hat g^h_t(p) = \frac{v^h_t\, g^h_t(p)}{\sum_{p'} v^h_t\, g^h_t(p') + \varepsilon},
\qquad \sigma = 1.8 .
\label{eq:footprint}
\end{equation}
The normalization makes the injected mass independent of $\sigma$, and an undetected hand contributes
nothing. Let $u^{(\ell)}_{t,p}$ denote the video token at latent frame $t$ and cell $p$ after block
$\ell$. Each injection block updates it as
\begin{equation}
u^{(\ell)}_{t,p} \;\leftarrow\; u^{(\ell)}_{t,p}
\;+\; \lambda \sum_{h\in\{\mathrm{L},\mathrm{R}\}} \hat g^h_t(p)\,
\Big( W^{(\ell)}\,\mathrm{LN}^{(\ell)}\!\big(e^h_t\big) + b^{(\ell)} \Big),
\label{eq:residual}
\end{equation}
where $\mathrm{LN}^{(\ell)}$ is a LayerNorm, $W^{(\ell)}\in\mathbb{R}^{d\times d}$ and $b^{(\ell)}$ are
a linear map and bias, and $\lambda$ is a scale fixed at one and stored as a buffer so that the pathway
can be disabled exactly at inference (\Cref{sec:exp:input}). We inject at blocks $\{0,6,12,18,24\}$.

\paragraph{Initialization and causal masking.} $W^{(\ell)}$ and $b^{(\ell)}$ are initialized to zero,
so the residual is zero at initialization while the projection receives gradient from the first
step (\Cref{app:training}). Setting $v^h_t = 0$ for every $t > m$ removes the residual from all
predicted latents, in training and evaluation.

\section{Experiments}
\label{sec:exp}

\begin{figure}[t]
\centering
\includegraphics[width=\textwidth]{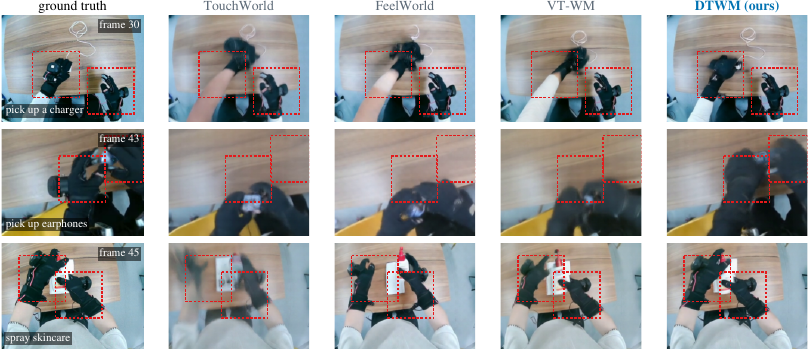}
\caption{\textbf{Comparison with other visual-tactile world models.} In these examples, \method{}
preserves hand shape and follows the ground-truth hand positions more closely than the other models.}
\label{fig:methods}
\end{figure}

We evaluate whether touch improves future-frame prediction and which parts of the tactile signal support
that improvement. We first compare \method{} with a matched vision-only baseline and three other visual-tactile world
models (\Cref{sec:exp:compare}), then remove information from the tactile input
(\Cref{sec:exp:input}). Finally, we examine visual quality and hand motion (\Cref{sec:exp:looks}), and
measure how the gains vary across training runs and prediction horizons (\Cref{sec:exp:horizon}).

\subsection{Experimental setup}
\label{sec:exp:setup}

\paragraph{Data and splits.} We use \corpus{} \citep{zhou2026touchanything}, which pairs egocentric
manipulation video with synchronized glove readings from both hands. Its \CorpusEpisodes{} episodes yield
\CorpusClips{} clips of $49$ frames, of which \SplitTrainClips{} are used for training. Our main evaluation
uses \FarmSubMethodSubNclips{} clips from recording sessions excluded from training. We divide these by
task: \FarmSubMethodSidNclips{} \emph{in-distribution} (ID) clips have the same, a near-duplicate, or a
related task in training; \FarmSubMethodSoodNclips{} \emph{out-of-distribution} (OOD) clips cover three
tasks with no close training counterpart (\Cref{tab:tasktiers}). For the three-run and horizon analyses,
we use two smaller splits: \emph{held-out episodes}, the first clip of each of
\FarmVisionOnlyOodNclips{} held-out episodes, and \emph{training episodes},
\FarmVisionOnlyIndomNclips{} reserved clips from the ends of training episodes. The former contains
three OOD clips; the latter is entirely ID. Neither split includes clips used for training
(\Cref{app:data}).

\paragraph{Prediction task.} Given RGB frames $0$ to $12$, the corresponding tactile readings and hand
skeletons, and a task instruction, the model predicts frames $13$ to $48$ ($1.2$ seconds). No future hand
pose, action, or tactile reading is provided. All $36$ future frames are generated in one sampling pass
with block-causal attention, in chunks covering frames $13$ to $28$, $29$ to $44$, and $45$ to $48$. We
score the full predicted interval and analyze shorter windows in \Cref{sec:exp:horizon}.

\paragraph{Metrics and uncertainty.} Our primary metric is full-frame LPIPS
\citep{zhang2018unreasonable}. We also report LPIPS$_{\text{hand}}$ on an enlarged crop around the
ground-truth hands, which usually includes the manipulated object. Both metrics are lower when better.
Hand silhouette and optical-flow metrics assess spatial overlap and motion (\Cref{app:handmetrics}).
PSNR and SSIM \citep{wang2004ssim}, together with an analysis of their sensitivity to blur, are reported
in \Cref{app:pixelmetrics}. Percentage reductions are relative to the comparison model, and the statistical procedure is described
in \Cref{app:protocol}.

\paragraph{Baselines.} The vision-only baseline uses the same architecture, parameter count, and
training recipe as \method{}, but its tactile readings are zeroed before the embedding during both
training and inference. Its tactile pathway remains trainable and retains the same injection sites and
hand locations, so the comparison tests the contribution of the sensor readings. We also compare with three
other visual-tactile world models, TouchWorld \citep{zhou2026touchworld}, VT-WM
\citep{higuera2026visuotactile}, and FeelWorld \citep{ma2026feelworld}, whose tactile conditioning mechanisms
we adapt to our backbone, glove data, and training recipe (\Cref{app:reimpl}). ViTacWorld \citep{huang2026vitacworld},
which requires optical tactile images, is not included.

\paragraph{Training.}\label{sec:method:training} We freeze the video autoencoder and text encoder, and
adapt the pretrained transformer with rank-$64$ LoRA \citep{hu2022lora}. The widened patch embedding,
the tactile embedding, and the LayerNorm and projection that add the tactile residual at five blocks are
trained at full rank. The flow-matching loss assigns
weights of $0$, $1$, and $0.2$ to observed latent frames, the first predicted chunk, and later frames,
respectively. A spatial weight emphasizes the hands; future ground-truth hand locations are used only to
weight the loss. All models train for one epoch over \SplitTrainClips{} clips using AdamW at
$3\times10^{-5}$ after a $100$-step warmup (\Cref{app:training}).

\subsection{Comparison with the vision-only baseline and other visual-tactile world models}
\label{sec:exp:compare}
\label{sec:exp:main}

\begin{figure}[t]
\centering
\includegraphics[width=\textwidth]{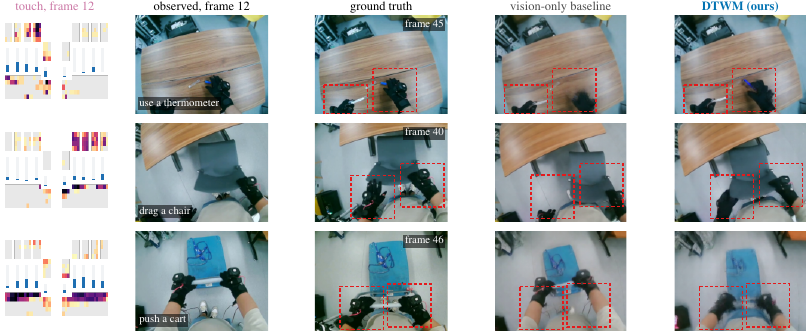}
\caption{\textbf{Examples on OOD tasks.} On tasks with no close counterpart in training, \method{}
preserves more hand detail and follows the ground-truth hand positions more closely than the vision-only
baseline in the examples shown.}
\label{fig:ood}
\end{figure}

\begin{table}[t]
\centering
\caption{\textbf{Prediction quality on the held-out set.} \method{} achieves the lowest full-frame and
hand-crop LPIPS on both task splits. Hand overlap (IoU) and flow direction agreement are measured over
all clips. All rows report one training run.}
\label{tab:compare}
\footnotesize
\setlength{\tabcolsep}{2.6pt}
\begin{tabular}{lcccccc}
\toprule
& \multicolumn{2}{c}{ID (2,728 clips)} & \multicolumn{2}{c}{OOD (272 clips)} & \multicolumn{2}{c}{Hand, all 3,000 clips} \\
\cmidrule(lr){2-3}\cmidrule(lr){4-5}\cmidrule(lr){6-7}
Model & LPIPS $\downarrow$ & LPIPS$_{\text{hand}}$ $\downarrow$ & LPIPS $\downarrow$ & LPIPS$_{\text{hand}}$ $\downarrow$ & IoU $\uparrow$ & flow dir. $\uparrow$ \\
\midrule
Vision-only baseline & 0.4243 & 0.3273 & 0.4882 & 0.3765 & 0.475 & 0.039 \\
TouchWorld \citep{zhou2026touchworld} & 0.4899 & 0.3979 & 0.5341 & 0.4204 & 0.376 & 0.031 \\
VT-WM \citep{higuera2026visuotactile} & 0.4237 & 0.3324 & 0.4854 & 0.3797 & 0.462 & 0.032 \\
FeelWorld \citep{ma2026feelworld} & 0.4488 & 0.3465 & 0.5071 & 0.3964 & 0.430 & 0.027 \\
\midrule
\method{} (ours) & \textbf{0.4157} & \textbf{0.3227} & \textbf{0.4774} & \textbf{0.3668} & \textbf{0.476} & \textbf{0.041} \\
\bottomrule
\end{tabular}

\end{table}

\method{} achieves the lowest LPIPS and LPIPS$_{\text{hand}}$ on both ID and OOD clips
(\Cref{tab:compare}). Relative to the vision-only baseline, full-frame LPIPS decreases by
\DconSubMainSidLpipsRel{} on ID clips and
\DconSubMainSoodLpipsRel{} on OOD clips. The same holds on OOD clips, which involve unseen tasks
and unseen objects.

The other visual-tactile world models receive the same glove measurements and differ only in how the
reading conditions the video model: TouchWorld renders it as image panels, VT-WM appends it as tokens,
and FeelWorld adds a gated global feature. \method{} instead adds each hand's reading at that hand's
image location, and it has the lowest perceptual error of the four designs. With the same signal and
training, how the tactile signal is fed to the model therefore affects prediction quality.

\subsection{What information in touch matters?}
\label{sec:exp:input}
\label{sec:exp:carry}

\begin{table}[t]
\centering
\caption{\textbf{Ablation of tactile input.} The full reading gives the lowest perceptual error.
Per-taxel contact retains contact locations and finger flexion but removes pressure magnitude; one bit
per hand retains only contact state. The pathway-disabled row evaluates the full-reading checkpoint
with its tactile residual switched off. Lower is better for all metrics.}
\label{tab:input}
\footnotesize
\setlength{\tabcolsep}{3pt}
\begin{tabular}{lcccc}
\toprule
& \multicolumn{2}{c}{ID (2,728 clips)} & \multicolumn{2}{c}{OOD (272 clips)} \\
\cmidrule(lr){2-3}\cmidrule(lr){4-5}
Tactile input & LPIPS & LPIPS$_{\text{hand}}$ & LPIPS & LPIPS$_{\text{hand}}$ \\
\midrule
zeroed reading (vision-only baseline) & 0.4243 & 0.3273 & 0.4882 & 0.3765 \\
pathway disabled at inference & 0.4208 & 0.3297 & 0.4838 & 0.3754 \\
\midrule
one contact bit per hand & 0.4307 & 0.3318 & 0.5014 & 0.3864 \\
per-taxel contact, no force & 0.4200 & 0.3246 & 0.4823 & 0.3745 \\
\method{} (ours, full reading) & \textbf{0.4157} & \textbf{0.3227} & \textbf{0.4774} & \textbf{0.3668} \\
\bottomrule
\end{tabular}

\end{table}

\Cref{tab:input} compares two coarser tactile inputs with the full reading, using one training run per
input and keeping the architecture and training recipe fixed. It also evaluates the full-reading model
with its tactile pathway disabled at inference.

\begin{wrapfigure}[19]{r}{0.42\textwidth}
\vspace{-12pt}
\centering
\includegraphics[width=0.42\textwidth]{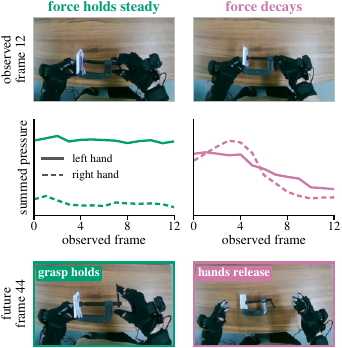}
\caption{\textbf{Touch history correlates with future manipulation.} The observed frames look alike, but the observed touch evolves differently, and so do the manipulations that follow.}
\label{fig:motivation}
\end{wrapfigure}
\paragraph{Pressure magnitude improves prediction.} Thresholding each pressure taxel keeps the spatial pattern
of contact and the finger-flexion channels but removes pressure magnitude. This increases full-frame and hand-crop LPIPS on both splits. Replacing the reading
with one contact bit per hand, which also discards where the hand is pressed and how its fingers are
flexed, further increases error and makes the model worse than the vision-only baseline. The model
therefore uses more than whether a hand is in contact.

\paragraph{Touch history correlates with future manipulation.} The observation records how pressure
changes over time. In the training corpus, a hand whose force holds steady rarely releases its grasp within
the next chunk, whereas a hand whose force rises or decays is \ForceIncVsSteady{} to
\ForceDecVsSteady{} times as likely to release it (\Cref{fig:motivation,app:contact}). The trend remains predictive
after accounting for the current force level. Binarizing the pressure removes both the force level and its
trend, and with them this cue to how the interaction will continue.

\paragraph{Some benefit persists without touch at inference.} Setting $\lambda=0$ in
\Cref{eq:residual} disables the tactile residual of the trained full-reading model. This worsens all
four perceptual scores relative to using the full reading, but full-frame LPIPS remains below the
vision-only baseline on both ID and OOD clips. We attribute this to training on paired touch and
video. Every training clip shows which visual appearance corresponds to which contact, where the hand
touches and how firmly, and the model learns this association, which remains useful when the tactile
pathway is off.

\subsection{Visual quality and hand motion}
\label{sec:exp:looks}

\paragraph{Visual detail.} The examples in \Cref{fig:id,fig:methods,fig:ood} show the improvement most
clearly around the hands. In the predictions of the other models, fingers can merge, hand outlines blur, and hands can
separate from the objects they hold. \method{} better preserves these details in the examples shown.
Across clips, its predictions also retain more high-frequency content than those of the vision-only
baseline, although both remain less sharp than the ground truth (\Cref{app:sharpness}). PSNR and SSIM
favor the vision-only baseline; a controlled blur analysis shows that these metrics can improve as
predictions lose detail, while LPIPS worsens (\Cref{app:pixelmetrics}). We therefore assess perceptual
quality alongside the hand-behavior metrics below.

\paragraph{Hand motion.} \method{} improves all reported hand-behavior metrics over the other visual-tactile
world models (\Cref{tab:radar:app}). Against the matched vision-only baseline, touch corrects how far
the hands move and how large they appear (\Cref{tab:hand:app}). The underestimation of optical-flow
magnitude falls from \HandMethodPVsControlPFullFlowMagGapB{} to
\HandMethodPVsControlPFullFlowMagGapA{}, and the silhouette area ratio moves closer to one.

\subsection{Variation across runs and prediction horizons}
\label{sec:exp:horizon}
\label{sec:exp:runs}

\begin{figure}[t]
\centering
\includegraphics[width=0.84\textwidth]{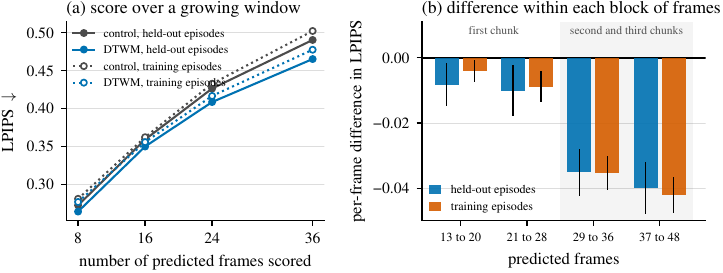}
\caption{\textbf{The LPIPS improvement grows over the prediction horizon.} (a) Cumulative scoring
windows: \method{} has lower LPIPS at every window, with a larger gap beyond the first chunk.
(b) Disjoint temporal segments: on held-out episodes, the average per-frame gap after the first chunk
is about \BlockRatioOod{} times that within it. Scores are averaged over three runs per model.}
\label{fig:horizon}
\end{figure}

\paragraph{Training runs.} Across three independently trained runs per model, \method{} has lower
mean LPIPS on both smaller evaluation splits (\Cref{tab:main}). It wins
\RunPairsOodNeg{} of \RunPairsOodN{} cross-model run comparisons on held-out episodes and
\RunPairsIndomNeg{} of \RunPairsIndomN{} on training episodes (\Cref{tab:perrun:app}). The advantage
therefore holds on average and in most pairings of individual runs.

\paragraph{Prediction horizon.} The LPIPS gap is small in the first predicted chunk and grows in the
later chunks on both splits (\Cref{fig:horizon,tab:curve:app}). On held-out episodes, the mean per-frame
improvement after the first chunk is about \BlockRatioOod{} times that within the first chunk. Because
the same generated videos are scored over each window, this pattern reflects where the benefit appears
within a prediction. It is consistent with touch resolving uncertainty about the observed interaction:
an error in that estimate barely affects the first predicted frames, but later chunks attend to earlier
ones and inherit their errors.

\section{Conclusion}
\label{sec:conclusion}

We presented \method{}, a world model that predicts egocentric human manipulation from the observed video
and the tactile signals of both hands. We finetune a pretrained video diffusion transformer and add each
hand's tactile signal to the video tokens at that hand's location through a zero-initialized, causally
masked residual. Compared with the vision-only baseline and existing visual-tactile world models,
\method{} better predicts how the hands move and appear under the same setting, and its advantage over
the vision-only baseline grows over the prediction horizon. Training with touch improves future-frame
prediction even when no touch is available at inference. Ablations show that the model benefits from the
magnitude and location of the force rather than from contact alone. Touch thus provides a direct signal of
the contact events that video struggles to capture, and it makes the predicted hand-object interaction
more accurate.

\paragraph{Limitations.} To our knowledge, the corpus we use is the only open dataset that pairs
egocentric video of everyday human manipulation with tactile signals from both hands, and at about
twenty hours it is small for video models. Hand locations come from an off-the-shelf
reconstruction that misses some frames. Finally, the model predicts human manipulation, so it cannot yet be placed directly
inside a robot planner or policy. Extending it to a robot embodiment is a natural next step.

\section*{Ethics statement}

This work uses recordings of people who wore instrumented gloves while carrying out everyday tasks in
homes, offices, shops, workshops and outdoor settings \citep{zhou2026touchanything}. The frames may show
private interiors, public spaces, personal belongings and, incidentally, other people, and the gloves
record the wearer's hands throughout. We collected no new recordings, used the corpus for research
purposes only, and do not redistribute raw video. The corpus authors state that the data, code and
benchmark will be released publicly \citep{zhou2026touchanything}; we will follow the license terms they
publish, and our own release will cover code, configurations, split lists and per-clip scores rather than
imagery. As with manipulation research in general, the method has potential dual uses. Improved video
prediction of manipulation can serve robot learning, teleoperation and assistive devices, and we are not
aware of a misuse specific to conditioning a video model on a body-worn sensor beyond those already
associated with video generation and robot learning.

\section*{Reproducibility statement}

The appendices describe the corpus and splits (\Cref{app:data}), the tactile channels (\Cref{app:sensor}),
the model and training recipe (\Cref{app:training}), the evaluation protocol and statistics
(\Cref{app:protocol}), and the complete results (\Cref{app:results}). We will release code, per-model
configurations, split lists and per-clip scores.

\section*{AI use statement}

We used an LLM-based coding assistant during this project and describe its use in the categories
requested by the conference policy.

\paragraph{Tasks for which it was used.} The assistant helped draft and revise the text of this paper,
including this statement. It assisted with literature search, in particular for the recent visuotactile
world models discussed in \Cref{sec:related}. It wrote and executed training and evaluation scripts, the
statistical analysis and the figure scripts, and it implemented and launched several of the ablation
models.

\paragraph{Tasks for which it was not used.} No part of the data is synthetic or generated by a model;
all data are the released \corpus{} recordings. The paper contains no mathematical result whose proof was
produced by an LLM.

\paragraph{Review of the assisted work.} Every score, difference and interval in the text and tables is
emitted by a script that reads the per-clip score files, so no reported figure can diverge from the data
underlying it. Descriptive statistics of the corpus and configuration constants are transcribed from the
pipeline's logs. Every bibliography entry was checked against its Crossref or arXiv record. Figures are
generated from held-out clips by released scripts, and the frames shown were selected by documented rules.
Statements about prior work were checked against the original papers rather than against summaries
produced by the assistant. We reviewed all assisted work and take full responsibility for the content of
this paper, including text, claims and artifacts produced with the help of generative AI.

\bibliography{refs}

\begin{thebibliography}{56}
\providecommand{\natexlab}[1]{#1}
\providecommand{\url}[1]{\texttt{#1}}
\expandafter\ifx\csname urlstyle\endcsname\relax
  \providecommand{\doi}[1]{doi: #1}\else
  \providecommand{\doi}{doi: \begingroup \urlstyle{rm}\Url}\fi

\bibitem[Black et~al.(2025)Black, Brown, Driess, Esmail, Equi, Finn, Fusai,
  Groom, Hausman, Ichter, et~al.]{black2025pi0}
Kevin Black, Noah Brown, Danny Driess, Adnan Esmail, Michael Equi, Chelsea
  Finn, Niccolo Fusai, Lachy Groom, Karol Hausman, Brian Ichter, et~al.
\newblock $\pi_0$: A vision-language-action flow model for general robot
  control.
\newblock In \emph{Robotics: Science and Systems XXI}, 2025.
\newblock \doi{10.15607/RSS.2025.XXI.010}.

\bibitem[Blattmann et~al.(2023{\natexlab{a}})Blattmann, Dockhorn, Kulal,
  Mendelevitch, Kilian, Lorenz, Levi, English, Voleti, Letts, Jampani, and
  Rombach]{blattmann2023stable}
Andreas Blattmann, Tim Dockhorn, Sumith Kulal, Daniel Mendelevitch, Maciej
  Kilian, Dominik Lorenz, Yam Levi, Zion English, Vikram Voleti, Adam Letts,
  Varun Jampani, and Robin Rombach.
\newblock {Stable Video Diffusion}: Scaling latent video diffusion models to
  large datasets.
\newblock \emph{arXiv preprint arXiv:2311.15127}, 2023{\natexlab{a}}.
\newblock \doi{10.48550/arXiv.2311.15127}.

\bibitem[Blattmann et~al.(2023{\natexlab{b}})Blattmann, Rombach, Ling,
  Dockhorn, Kim, Fidler, and Kreis]{blattmann2023align}
Andreas Blattmann, Robin Rombach, Huan Ling, Tim Dockhorn, Seung~Wook Kim,
  Sanja Fidler, and Karsten Kreis.
\newblock Align your latents: High-resolution video synthesis with latent
  diffusion models.
\newblock In \emph{2023 IEEE/CVF Conference on Computer Vision and Pattern
  Recognition (CVPR)}, pp.\  22563--22575. IEEE, 2023{\natexlab{b}}.
\newblock \doi{10.1109/CVPR52729.2023.02161}.

\bibitem[Bruce et~al.(2024)Bruce, Dennis, Edwards, Parker-Holder, Shi, Hughes,
  Lai, Mavalankar, Steigerwald, Apps, Aytar, Bechtle, Behbahani, Chan, Heess,
  Gonzalez, Osindero, Ozair, Reed, Zhang, Zolna, Clune, Freitas, Singh, and
  Rockt{\"a}schel]{bruce2024genie}
Jake Bruce, Michael~D Dennis, Ashley Edwards, Jack Parker-Holder, Yuge Shi,
  Edward Hughes, Matthew Lai, Aditi Mavalankar, Richie Steigerwald, Chris Apps,
  Yusuf Aytar, Sarah Maria~Elisabeth Bechtle, Feryal Behbahani, Stephanie~C.Y.
  Chan, Nicolas Heess, Lucy Gonzalez, Simon Osindero, Sherjil Ozair, Scott
  Reed, Jingwei Zhang, Konrad Zolna, Jeff Clune, Nando~De Freitas, Satinder
  Singh, and Tim Rockt{\"a}schel.
\newblock Genie: Generative interactive environments.
\newblock In \emph{Proceedings of the 41st International Conference on Machine
  Learning}, volume 235 of \emph{Proceedings of Machine Learning Research},
  pp.\  4603--4623. PMLR, 2024.
\newblock URL \url{https://proceedings.mlr.press/v235/bruce24a.html}.

\bibitem[Calandra et~al.(2018)Calandra, Owens, Jayaraman, Lin, Yuan, Malik,
  Adelson, and Levine]{calandra2018regrasp}
Roberto Calandra, Andrew Owens, Dinesh Jayaraman, Justin Lin, Wenzhen Yuan,
  Jitendra Malik, Edward~H. Adelson, and Sergey Levine.
\newblock More than a feeling: Learning to grasp and regrasp using vision and
  touch.
\newblock \emph{IEEE Robotics and Automation Letters}, 3\penalty0 (4):\penalty0
  3300--3307, 2018.
\newblock \doi{10.1109/LRA.2018.2852779}.

\bibitem[Chen et~al.(2024)Chen, {Mart{\'\i} Mons{\'o}}, Du, Simchowitz,
  Tedrake, and Sitzmann]{chen2024diffusionforcing}
Boyuan Chen, Diego {Mart{\'\i} Mons{\'o}}, Yilun Du, Max Simchowitz, Russ
  Tedrake, and Vincent Sitzmann.
\newblock {Diffusion Forcing}: Next-token prediction meets full-sequence
  diffusion.
\newblock In \emph{Advances in Neural Information Processing Systems 37: Annual
  Conference on Neural Information Processing Systems 2024, NeurIPS 2024,
  Vancouver, BC, Canada, December 10 - 15, 2024}, pp.\  24081--24125, 2024.
\newblock \doi{10.52202/079017-0759}.
\newblock URL
  \url{http://papers.nips.cc/paper\_files/paper/2024/hash/2aee1c4159e48407d68fe16ae8e6e49e-Abstract-Conference.html}.

\bibitem[Chen et~al.(2026)Chen, Liu, Li, Zeng, Zhu, Cong, Hong, Yang, Tu, Wang,
  Ivanovic, Pavone, Wang, Zhou, and Fan]{chen2026crashtwin}
Nuo Chen, Lulin Liu, Zihao Li, Ziyao Zeng, Zihao Zhu, Wenyan Cong, Junyuan
  Hong, Yunhao Yang, Zhengzhong Tu, Yan Wang, Boris Ivanovic, Marco Pavone,
  Zhangyang Wang, Yang Zhou, and Zhiwen Fan.
\newblock A physics-grounded benchmark for multi-agent dynamics in world
  models.
\newblock In \emph{Computer Vision -- {ECCV} 2026: 19th European Conference,
  Malm{\"o}, Sweden, September 8--12, 2026, Proceedings, Part {LXXIV}}, Lecture
  Notes in Computer Science, pp.\  547--565. Springer Nature Switzerland, 2026.
\newblock \doi{10.1007/978-3-032-37029-7_32}.

\bibitem[Finn \& Levine(2017)Finn and Levine]{finn2017visualforesight}
Chelsea Finn and Sergey Levine.
\newblock Deep visual foresight for planning robot motion.
\newblock In \emph{2017 IEEE International Conference on Robotics and
  Automation (ICRA)}, pp.\  2786--2793. IEEE, 2017.
\newblock \doi{10.1109/ICRA.2017.7989324}.

\bibitem[Fu et~al.(2024)Fu, Datta, Huang, Panitch, Drake, Ortiz, Mukadam,
  Lambeta, Calandra, and Goldberg]{fu2024touch}
Letian Fu, Gaurav Datta, Huang Huang, William Chung-Ho Panitch, Jaimyn Drake,
  Joseph Ortiz, Mustafa Mukadam, Mike Lambeta, Roberto Calandra, and Ken
  Goldberg.
\newblock A touch, vision, and language dataset for multimodal alignment.
\newblock In \emph{Proceedings of the 41st International Conference on Machine
  Learning}, volume 235 of \emph{Proceedings of Machine Learning Research},
  pp.\  14080--14101. PMLR, 2024.
\newblock URL \url{https://proceedings.mlr.press/v235/fu24b.html}.

\bibitem[Ha \& Schmidhuber(2018)Ha and Schmidhuber]{ha2018recurrent}
David Ha and J{\"{u}}rgen Schmidhuber.
\newblock Recurrent world models facilitate policy evolution.
\newblock In Samy Bengio, Hanna~M. Wallach, Hugo Larochelle, Kristen Grauman,
  Nicol{\`{o}} Cesa{-}Bianchi, and Roman Garnett (eds.), \emph{Advances in
  Neural Information Processing Systems 31: Annual Conference on Neural
  Information Processing Systems 2018, NeurIPS 2018, December 3-8, 2018,
  Montr{\'{e}}al, Canada}, pp.\  2455--2467, 2018.
\newblock URL
  \url{https://proceedings.neurips.cc/paper/2018/hash/2de5d16682c3c35007e4e92982f1a2ba-Abstract.html}.

\bibitem[Hafner et~al.(2020)Hafner, Lillicrap, Ba, and
  Norouzi]{hafner2020dream}
Danijar Hafner, Timothy Lillicrap, Jimmy Ba, and Mohammad Norouzi.
\newblock Dream to control: Learning behaviors by latent imagination.
\newblock In \emph{International Conference on Learning Representations}, 2020.
\newblock URL \url{https://openreview.net/forum?id=S1lOTC4tDS}.

\bibitem[Higuera et~al.(2026)Higuera, Arnaud, Boots, Mukadam, Hogan, and
  Meier]{higuera2026visuotactile}
Carolina Higuera, Sergio Arnaud, Byron Boots, Mustafa Mukadam, Francois~Robert
  Hogan, and Franziska Meier.
\newblock Visuo-tactile world models.
\newblock arXiv preprint arXiv:2602.06001, 2026.

\bibitem[Hu et~al.(2022)Hu, Shen, Wallis, Allen-Zhu, Li, Wang, Wang, and
  Chen]{hu2022lora}
Edward~J. Hu, Yelong Shen, Phillip Wallis, Zeyuan Allen-Zhu, Yuanzhi Li, Shean
  Wang, Lu~Wang, and Weizhu Chen.
\newblock {LoRA}: Low-rank adaptation of large language models.
\newblock In \emph{The Tenth International Conference on Learning
  Representations, {ICLR} 2022, Virtual Event, April 25-29, 2022}.
  OpenReview.net, 2022.
\newblock URL \url{https://openreview.net/forum?id=nZeVKeeFYf9}.

\bibitem[Hu et~al.(2025)Hu, Guo, Wang, Chen, Wang, Zhang, Sreenath, Lu, and
  Chen]{hu2025vpp}
Yucheng Hu, Yanjiang Guo, Pengchao Wang, Xiaoyu Chen, Yen-Jen Wang, Jianke
  Zhang, Koushil Sreenath, Chaochao Lu, and Jianyu Chen.
\newblock Video prediction policy: A generalist robot policy with predictive
  visual representations.
\newblock In \emph{Proceedings of the 42nd International Conference on Machine
  Learning}, volume 267 of \emph{Proceedings of Machine Learning Research},
  pp.\  24328--24346. PMLR, 2025.
\newblock URL \url{https://proceedings.mlr.press/v267/hu25g.html}.

\bibitem[Huang et~al.(2025)Huang, Wang, Yang, Luo, and Li]{huang2024vitac}
Binghao Huang, Yixuan Wang, Xinyi Yang, Yiyue Luo, and Yunzhu Li.
\newblock {3D-ViTac}: Learning fine-grained manipulation with visuo-tactile
  sensing.
\newblock In Pulkit Agrawal, Oliver Kroemer, and Wolfram Burgard (eds.),
  \emph{Proceedings of The 8th Conference on Robot Learning}, volume 270 of
  \emph{Proceedings of Machine Learning Research}, pp.\  2557--2578. PMLR,
  2025.
\newblock URL \url{https://proceedings.mlr.press/v270/huang25e.html}.

\bibitem[Huang et~al.(2026)Huang, Sang, Lu, Ni, Wu, Guo, Shi, and
  Wang]{huang2026vitacworld}
Yunao Huang, Shiyu Sang, Haotao Lu, Suting Ni, Shijie Wu, Ziyang Guo, Ye~Shi,
  and Jingya Wang.
\newblock {ViTacWorld}: Scaling visuo-tactile world models for contact-rich
  robot manipulation.
\newblock arXiv preprint arXiv:2607.22530, 2026.

\bibitem[Jang et~al.(2025)Jang, Ye, Lin, Xiang, Bjorck, Fang, Hu, Huang,
  Kundalia, Lin, et~al.]{jang2025dreamgen}
Joel Jang, Seonghyeon Ye, Zongyu Lin, Jiannan Xiang, Johan Bjorck, Yu~Fang,
  Fengyuan Hu, Spencer Huang, Kaushil Kundalia, Yen-Chen Lin, et~al.
\newblock {DreamGen}: Unlocking generalization in robot learning through video
  world models.
\newblock In \emph{Proceedings of The 9th Conference on Robot Learning}, volume
  305 of \emph{Proceedings of Machine Learning Research}, pp.\  5170--5194.
  PMLR, 2025.
\newblock URL \url{https://proceedings.mlr.press/v305/jang25a.html}.

\bibitem[Kim et~al.(2025)Kim, Pertsch, Karamcheti, Xiao, Balakrishna, Nair,
  Rafailov, Foster, Sanketi, Vuong, Kollar, Burchfiel, Tedrake, Sadigh, Levine,
  Liang, and Finn]{kim2024openvla}
Moo~Jin Kim, Karl Pertsch, Siddharth Karamcheti, Ted Xiao, Ashwin Balakrishna,
  Suraj Nair, Rafael Rafailov, Ethan~P Foster, Pannag~R Sanketi, Quan Vuong,
  Thomas Kollar, Benjamin Burchfiel, Russ Tedrake, Dorsa Sadigh, Sergey Levine,
  Percy Liang, and Chelsea Finn.
\newblock {OpenVLA}: An open-source vision-language-action model.
\newblock In \emph{Proceedings of The 8th Conference on Robot Learning}, volume
  270 of \emph{Proceedings of Machine Learning Research}, pp.\  2679--2713.
  PMLR, 2025.
\newblock URL \url{https://proceedings.mlr.press/v270/kim25c.html}.

\bibitem[Lambeta et~al.(2020)Lambeta, Chou, Tian, Yang, Maloon, Most, Stroud,
  Santos, Byagowi, Kammerer, Jayaraman, and Calandra]{lambeta2020digit}
Mike Lambeta, Po-Wei Chou, Stephen Tian, Brian Yang, Benjamin Maloon,
  Victoria~Rose Most, Dave Stroud, Raymond Santos, Ahmad Byagowi, Gregg
  Kammerer, Dinesh Jayaraman, and Roberto Calandra.
\newblock {DIGIT}: A novel design for a low-cost compact high-resolution
  tactile sensor with application to in-hand manipulation.
\newblock \emph{IEEE Robotics and Automation Letters}, 5\penalty0 (3):\penalty0
  3838--3845, 2020.
\newblock \doi{10.1109/LRA.2020.2977257}.

\bibitem[Lee et~al.(2019)Lee, Zhu, Srinivasan, Shah, Savarese, Fei-Fei, Garg,
  and Bohg]{lee2019visiontouch}
Michelle~A. Lee, Yuke Zhu, Krishnan Srinivasan, Parth Shah, Silvio Savarese,
  Li~Fei-Fei, Animesh Garg, and Jeannette Bohg.
\newblock Making sense of vision and touch: Self-supervised learning of
  multimodal representations for contact-rich tasks.
\newblock In \emph{2019 International Conference on Robotics and Automation
  (ICRA)}, pp.\  8943--8950. IEEE, 2019.
\newblock \doi{10.1109/icra.2019.8793485}.

\bibitem[Lipman et~al.(2023)Lipman, Chen, Ben-Hamu, Nickel, and
  Le]{lipman2023flow}
Yaron Lipman, Ricky T.~Q. Chen, Heli Ben-Hamu, Maximilian Nickel, and Matthew
  Le.
\newblock Flow matching for generative modeling.
\newblock In \emph{The Eleventh International Conference on Learning
  Representations, {ICLR} 2023, Kigali, Rwanda, May 1-5, 2023}. OpenReview.net,
  2023.
\newblock URL \url{https://openreview.net/forum?id=PqvMRDCJT9t}.

\bibitem[Liu et~al.(2026)Liu, Chen, Wang, Liu, Cong, Hu, Ivanovic, Wang, Zeng,
  Gong, Zhou, Xiong, Wang, Wang, Shi, Zhang, Pavone, and
  Fan]{liu2026openlongtail}
Lulin Liu, Nuo Chen, Yan Wang, Bangya Liu, Wenyan Cong, Hezhen Hu, Boris
  Ivanovic, Hao Wang, Ziyao Zeng, Xinyu Gong, Yang Zhou, Zixiang Xiong, Dilin
  Wang, Zhangyang Wang, Weisong Shi, Ruohan Zhang, Marco Pavone, and Zhiwen
  Fan.
\newblock {OpenLongTail}: Generative scaling of long-tail driving data.
\newblock arXiv preprint arXiv:2607.09655, 2026.

\bibitem[Liu et~al.(2023)Liu, Gong, and Liu]{liu2023rectifiedflow}
Xingchao Liu, Chengyue Gong, and Qiang Liu.
\newblock Flow straight and fast: Learning to generate and transfer data with
  rectified flow.
\newblock In \emph{The Eleventh International Conference on Learning
  Representations, {ICLR} 2023, Kigali, Rwanda, May 1-5, 2023}. OpenReview.net,
  2023.
\newblock URL \url{https://openreview.net/forum?id=XVjTT1nw5z}.

\bibitem[Loshchilov \& Hutter(2019)Loshchilov and
  Hutter]{loshchilov2019decoupled}
Ilya Loshchilov and Frank Hutter.
\newblock Decoupled weight decay regularization.
\newblock In \emph{7th International Conference on Learning Representations,
  {ICLR} 2019, New Orleans, LA, USA, May 6-9, 2019}. OpenReview.net, 2019.
\newblock URL \url{https://openreview.net/forum?id=Bkg6RiCqY7}.

\bibitem[Luo et~al.(2026)Luo, Huang, Xu, Li, Jiao, and Xiao]{luo2026blind}
Shengcheng Luo, Xiyan Huang, Zhe Xu, Wanlin Li, Ziyuan Jiao, and Chenxi Xiao.
\newblock Blind dexterous grasping via {Real2Sim2Real} tactile policy learning.
\newblock arXiv preprint arXiv:2606.11767, 2026.

\bibitem[Ma et~al.(2026)Ma, Zhang, Xue, Cai, Yao, Cui, and
  Wang]{ma2026feelworld}
Wenxuan Ma, Chaofan Zhang, Chao Xue, Yinghao Cai, Guocai Yao, Shaowei Cui, and
  Shuo Wang.
\newblock {FeelWorld}: Visuo-tactile world model for hierarchical contact
  prediction and planning.
\newblock arXiv preprint arXiv:2607.24267, 2026.

\bibitem[Mou et~al.(2024)Mou, Wang, Xie, Wu, Zhang, Qi, and
  Shan]{mou2024t2iadapter}
Chong Mou, Xintao Wang, Liangbin Xie, Yanze Wu, Jian Zhang, Zhongang Qi, and
  Ying Shan.
\newblock {T2I-Adapter}: Learning adapters to dig out more controllable ability
  for text-to-image diffusion models.
\newblock In \emph{Proceedings of the AAAI Conference on Artificial
  Intelligence}, volume~38, pp.\  4296--4304. Association for the Advancement
  of Artificial Intelligence (AAAI), 2024.
\newblock \doi{10.1609/aaai.v38i5.28226}.

\bibitem[{NVIDIA} et~al.(2025){NVIDIA}, Agarwal, Ali, Bala, Balaji, Barker,
  Cai, Chattopadhyay, Chen, Cui, Ding, et~al.]{agarwal2025cosmos}
{NVIDIA}, Niket Agarwal, Arslan Ali, Maciej Bala, Yogesh Balaji, Erik Barker,
  Tiffany Cai, Prithvijit Chattopadhyay, Yongxin Chen, Yin Cui, Yifan Ding,
  et~al.
\newblock Cosmos world foundation model platform for physical {AI}.
\newblock \emph{arXiv preprint arXiv:2501.03575}, 2025.

\bibitem[Oh et~al.(2015)Oh, Guo, Lee, Lewis, and
  Singh]{oh2015actionconditional}
Junhyuk Oh, Xiaoxiao Guo, Honglak Lee, Richard~L. Lewis, and Satinder Singh.
\newblock Action-conditional video prediction using deep networks in {Atari}
  games.
\newblock In \emph{Advances in Neural Information Processing Systems 28 (NIPS
  2015)}, 2015.

\bibitem[Peebles \& Xie(2023)Peebles and Xie]{peebles2023scalable}
William Peebles and Saining Xie.
\newblock Scalable diffusion models with transformers.
\newblock In \emph{2023 IEEE/CVF International Conference on Computer Vision
  (ICCV)}, pp.\  4172--4182. IEEE, 2023.
\newblock \doi{10.1109/ICCV51070.2023.00387}.

\bibitem[Potamias et~al.(2025)Potamias, Zhang, Deng, and
  Zafeiriou]{potamias2025wilor}
Rolandos~Alexandros Potamias, Jinglei Zhang, Jiankang Deng, and Stefanos
  Zafeiriou.
\newblock {WiLoR}: End-to-end {3D} hand localization and reconstruction
  in-the-wild.
\newblock In \emph{2025 IEEE/CVF Conference on Computer Vision and Pattern
  Recognition (CVPR)}, pp.\  12242--12254. IEEE, 2025.
\newblock \doi{10.1109/CVPR52734.2025.01143}.

\bibitem[Quevedo et~al.(2026)Quevedo, Sharma, Sun, Suryavanshi, Liang, and
  Yang]{quevedo2025worldgym}
Julian Quevedo, Ansh~Kumar Sharma, Yixiang Sun, Varad Suryavanshi, Percy Liang,
  and Sherry Yang.
\newblock {WorldGym}: World model as an environment for policy evaluation.
\newblock In \emph{International Conference on Learning Representations}, 2026.
\newblock URL \url{https://openreview.net/forum?id=hidBHy1CAw}.

\bibitem[Rajbhandari et~al.(2020)Rajbhandari, Rasley, Ruwase, and
  He]{rajbhandari2020zero}
Samyam Rajbhandari, Jeff Rasley, Olatunji Ruwase, and Yuxiong He.
\newblock {ZeRO}: Memory optimizations toward training trillion parameter
  models.
\newblock In \emph{Proceedings of the International Conference for High
  Performance Computing, Networking, Storage and Analysis (SC)}, pp.\  1--16,
  2020.
\newblock \doi{10.1109/SC41405.2020.00024}.

\bibitem[Rombach et~al.(2022)Rombach, Blattmann, Lorenz, Esser, and
  Ommer]{rombach2022latent}
Robin Rombach, Andreas Blattmann, Dominik Lorenz, Patrick Esser, and Bj{\"o}rn
  Ommer.
\newblock High-resolution image synthesis with latent diffusion models.
\newblock In \emph{2022 IEEE/CVF Conference on Computer Vision and Pattern
  Recognition (CVPR)}, pp.\  10674--10685, 2022.
\newblock \doi{10.1109/CVPR52688.2022.01042}.

\bibitem[Sun et~al.(2026{\natexlab{a}})Sun, Liang, Zeng, Wang, Zhang, Sun, Li,
  and Rakita]{sun2026decoupling}
Xiatao Sun, Chen Liang, Ziyao Zeng, Qian Wang, Haoyang Zhang, Yue Sun, Qiucheng
  Li, and Daniel Rakita.
\newblock Decoupling vision, language, and action for efficient multi-task
  robot policies.
\newblock arXiv preprint arXiv:2609.18374, 2026{\natexlab{a}}.

\bibitem[Sun et~al.(2026{\natexlab{b}})Sun, Zhuang, Negrete, Coldea, Liang,
  Zhang, Liu, Zeng, Li, Wang, Miao, and Rakita]{sun2026foveated}
Xiatao Sun, Yuan Zhuang, Mateo Sanchez~Lopez Negrete, Matei-Victor Coldea, Chen
  Liang, Haoyang Zhang, Che Liu, Ziyao Zeng, Shawn Li, Qian Wang, Fei Miao, and
  Daniel Rakita.
\newblock Artificial foveated perception for mitigating shortcut learning in
  robotic foundation models.
\newblock arXiv preprint arXiv:2607.10655, 2026{\natexlab{b}}.

\bibitem[Sundaram et~al.(2019)Sundaram, Kellnhofer, Li, Zhu, Torralba, and
  Matusik]{sundaram2019tactileglove}
Subramanian Sundaram, Petr Kellnhofer, Yunzhu Li, Jun-Yan Zhu, Antonio
  Torralba, and Wojciech Matusik.
\newblock Learning the signatures of the human grasp using a scalable tactile
  glove.
\newblock \emph{Nature}, 569\penalty0 (7758):\penalty0 698--702, 2019.
\newblock \doi{10.1038/s41586-019-1234-z}.

\bibitem[Tu et~al.(2026)Tu, Yang, Ma, Yu, Zeng, Wu, Zhao, Tao, Zhang, Qian, and
  Wong]{tu2026unitac}
Jiahang Tu, Fengyu Yang, Chenyang Ma, Xihang Yu, Ziyao Zeng, Shaokai Wu, Hanbin
  Zhao, Zhi Tao, Chao Zhang, Hui Qian, and Alex Wong.
\newblock {UniTac}: A unified multimodal model for cross-sensor tactile
  understanding and generation.
\newblock In \emph{European Conference on Computer Vision (ECCV)}, 2026.

\bibitem[Wan et~al.(2025)Wan, Wang, Ai, Wen, Mao, Xie, Chen, Yu, Zhao, Yang,
  Zeng, Wang, Zhang, Zhou, Wang, Chen, Zhu, Zhao, Yan, Huang, Feng, Zhang, Li,
  Wu, Chu, Feng, Zhang, Sun, Fang, Wang, Gui, Weng, Shen, Lin, Wang, Wang,
  Zhou, Wang, Shen, Yu, Shi, Huang, Xu, Kou, Lv, Li, Liu, Wang, Zhang, Huang,
  Li, Wu, Liu, Pan, Zheng, Hong, Shi, Feng, Jiang, Han, Wu, and
  Liu]{wan2025video}
Team Wan, Ang Wang, Baole Ai, Bin Wen, Chaojie Mao, Chen-Wei Xie, Di~Chen,
  Feiwu Yu, Haiming Zhao, Jianxiao Yang, Jianyuan Zeng, Jiayu Wang, Jingfeng
  Zhang, Jingren Zhou, Jinkai Wang, Jixuan Chen, Kai Zhu, Kang Zhao, Keyu Yan,
  Lianghua Huang, Mengyang Feng, Ningyi Zhang, Pandeng Li, Pingyu Wu, Ruihang
  Chu, Ruili Feng, Shiwei Zhang, Siyang Sun, Tao Fang, Tianxing Wang, Tianyi
  Gui, Tingyu Weng, Tong Shen, Wei Lin, Wei Wang, Wei Wang, Wenmeng Zhou, Wente
  Wang, Wenting Shen, Wenyuan Yu, Xianzhong Shi, Xiaoming Huang, Xin Xu, Yan
  Kou, Yangyu Lv, Yifei Li, Yijing Liu, Yiming Wang, Yingya Zhang, Yitong
  Huang, Yong Li, You Wu, Yu~Liu, Yulin Pan, Yun Zheng, Yuntao Hong, Yupeng
  Shi, Yutong Feng, Zeyinzi Jiang, Zhen Han, Zhi-Fan Wu, and Ziyu Liu.
\newblock {Wan}: Open and advanced large-scale video generative models.
\newblock arXiv preprint arXiv:2503.20314, 2025.
\newblock URL \url{https://arxiv.org/abs/2503.20314}.

\bibitem[Wang et~al.(2020)Wang, Wang, Romero, Veiga, and
  Adelson]{wang2020swingbot}
Chen Wang, Shaoxiong Wang, Branden Romero, Filipe Veiga, and Edward Adelson.
\newblock {SwingBot}: Learning physical features from in-hand tactile
  exploration for dynamic swing-up manipulation.
\newblock In \emph{2020 IEEE/RSJ International Conference on Intelligent Robots
  and Systems (IROS)}, pp.\  5633--5640. IEEE, 2020.
\newblock \doi{10.1109/IROS45743.2020.9341006}.

\bibitem[Wang et~al.(2004)Wang, Bovik, Sheikh, and Simoncelli]{wang2004ssim}
Zhou Wang, Alan~C. Bovik, Hamid~R. Sheikh, and Eero~P. Simoncelli.
\newblock Image quality assessment: from error visibility to structural
  similarity.
\newblock \emph{{IEEE} Transactions on Image Processing}, 13\penalty0
  (4):\penalty0 600--612, 2004.
\newblock \doi{10.1109/TIP.2003.819861}.

\bibitem[Wu et~al.(2024)Wu, Jing, Cheang, Chen, Xu, Li, Liu, Li, and
  Kong]{wu2024gr1}
Hongtao Wu, Ya~Jing, Chilam Cheang, Guangzeng Chen, Jiafeng Xu, Xinghang Li,
  Minghuan Liu, Hang Li, and Tao Kong.
\newblock Unleashing large-scale video generative pre-training for visual robot
  manipulation.
\newblock In \emph{International Conference on Learning Representations}, 2024.

\bibitem[Yang et~al.(2024{\natexlab{a}})Yang, Feng, Chen, Park, Wang, Dou,
  Zeng, Chen, Gangopadhyay, Owens, and Wong]{yang2024unitouch}
Fengyu Yang, Chao Feng, Ziyang Chen, Hyoungseob Park, Daniel Wang, Yiming Dou,
  Ziyao Zeng, Xien Chen, Rit Gangopadhyay, Andrew Owens, and Alex Wong.
\newblock Binding touch to everything: Learning unified multimodal tactile
  representations.
\newblock In \emph{2024 IEEE/CVF Conference on Computer Vision and Pattern
  Recognition (CVPR)}, pp.\  26330--26343, 2024{\natexlab{a}}.
\newblock \doi{10.1109/CVPR52733.2024.02488}.

\bibitem[Yang et~al.(2024{\natexlab{b}})Yang, Du, Ghasemipour, Tompson,
  Kaelbling, Schuurmans, and Abbeel]{yang2024unisim}
Sherry Yang, Yilun Du, Seyed Kamyar~Seyed Ghasemipour, Jonathan Tompson,
  Leslie~Pack Kaelbling, Dale Schuurmans, and Pieter Abbeel.
\newblock Learning interactive real-world simulators.
\newblock In \emph{International Conference on Learning Representations},
  2024{\natexlab{b}}.

\bibitem[Ye et~al.(2023)Ye, Zhang, Liu, Han, and Yang]{ye2023ipadapter}
Hu~Ye, Jun Zhang, Sibo Liu, Xiao Han, and Wei Yang.
\newblock {IP-Adapter}: Text compatible image prompt adapter for text-to-image
  diffusion models.
\newblock arXiv preprint arXiv:2308.06721, 2023.

\bibitem[Yin et~al.(2025)Yin, Zhang, Zhang, Freeman, Durand, Shechtman, and
  Huang]{yin2025causvid}
Tianwei Yin, Qiang Zhang, Richard Zhang, William~T. Freeman, Fr{\'e}do Durand,
  Eli Shechtman, and Xun Huang.
\newblock From slow bidirectional to fast autoregressive video diffusion
  models.
\newblock In \emph{2025 IEEE/CVF Conference on Computer Vision and Pattern
  Recognition (CVPR)}, pp.\  22963--22974, Nashville, TN, USA, 2025. IEEE.
\newblock \doi{10.1109/cvpr52734.2025.02138}.

\bibitem[Yuan et~al.(2017)Yuan, Dong, and Adelson]{yuan2017gelsight}
Wenzhen Yuan, Siyuan Dong, and Edward~H. Adelson.
\newblock {GelSight}: High-resolution robot tactile sensors for estimating
  geometry and force.
\newblock \emph{Sensors}, 17\penalty0 (12):\penalty0 2762, 2017.
\newblock ISSN 1424-8220.
\newblock \doi{10.3390/s17122762}.

\bibitem[Zeng et~al.(2025)Zeng, Ni, Liu, and Wong]{zeng2025coffee}
Ziyao Zeng, Jingcheng Ni, Ruyi Liu, and Alex Wong.
\newblock {Coffee}: Controllable diffusion fine-tuning.
\newblock arXiv preprint arXiv:2511.14113, 2025.

\bibitem[Zeng et~al.(2026)Zeng, Ni, Wang, Rim, Chung, Yang, Hong, and
  Wong]{zeng2026iris}
Ziyao Zeng, Jingcheng Ni, Daniel Wang, Patrick Rim, Younjoon Chung, Fengyu
  Yang, Byung-Woo Hong, and Alex Wong.
\newblock {Iris}: Integrating language into diffusion-based monocular depth
  estimation.
\newblock In \emph{Proceedings of the IEEE/CVF Conference on Computer Vision
  and Pattern Recognition (CVPR)}, pp.\  34193--34205, 2026.

\bibitem[Zhang et~al.(2023)Zhang, Rao, and Agrawala]{zhang2023controlnet}
Lvmin Zhang, Anyi Rao, and Maneesh Agrawala.
\newblock Adding conditional control to text-to-image diffusion models.
\newblock In \emph{2023 IEEE/CVF International Conference on Computer Vision
  (ICCV)}, pp.\  3813--3824. IEEE, 2023.
\newblock \doi{10.1109/ICCV51070.2023.00355}.

\bibitem[Zhang et~al.(2024)Zhang, Han, Liu, Zhou, Lu, Qiao, Li, and
  Gao]{zhang2024llamaadapter}
Renrui Zhang, Jiaming Han, Chris Liu, Aojun Zhou, Pan Lu, Yu~Qiao, Hongsheng
  Li, and Peng Gao.
\newblock {LLaMA-Adapter}: Efficient fine-tuning of large language models with
  zero-initialized attention.
\newblock In \emph{The Twelfth International Conference on Learning
  Representations, {ICLR} 2024, Vienna, Austria, May 7-11, 2024}.
  OpenReview.net, 2024.
\newblock URL \url{https://openreview.net/forum?id=d4UiXAHN2W}.

\bibitem[Zhang et~al.(2018)Zhang, Isola, Efros, Shechtman, and
  Wang]{zhang2018unreasonable}
Richard Zhang, Phillip Isola, Alexei~A. Efros, Eli Shechtman, and Oliver Wang.
\newblock The unreasonable effectiveness of deep features as a perceptual
  metric.
\newblock In \emph{2018 IEEE/CVF Conference on Computer Vision and Pattern
  Recognition}, pp.\  586--595. IEEE, 2018.
\newblock \doi{10.1109/cvpr.2018.00068}.

\bibitem[Zhang et~al.(2026)Zhang, Zhou, Zhang, Desai, Amosa, Soleymanzadeh,
  Lei, Zhou, Zheng, and She]{zhang2026contactworld}
Zhiyuan Zhang, Pokuang Zhou, Kaidi Zhang, Adeesh Desai, Temitope Amosa, Davood
  Soleymanzadeh, Jiuzhou Lei, Yuhao Zhou, Minghui Zheng, and Yu~She.
\newblock {ContactWorld}: What representations matter for vision-tactile latent
  world models in contact-rich manipulation.
\newblock arXiv preprint arXiv:2606.13877, 2026.

\bibitem[Zhao et~al.(2025)Zhao, Kwon, Streli, Pollefeys, and
  Holz]{zhao2025egopressure}
Yiming Zhao, Taein Kwon, Paul Streli, Marc Pollefeys, and Christian Holz.
\newblock {EgoPressure}: A dataset for hand pressure and pose estimation in
  egocentric vision.
\newblock In \emph{{IEEE/CVF} Conference on Computer Vision and Pattern
  Recognition, {CVPR} 2025, Nashville, TN, USA, June 11-15, 2025}, pp.\
  27727--27738. Computer Vision Foundation / {IEEE}, 2025.
\newblock \doi{10.1109/CVPR52734.2025.02582}.

\bibitem[Zhou et~al.(2026{\natexlab{a}})Zhou, Gao, Hong, Liu, Zhang, Dai, Zhen,
  Lyu, Wu, Mao, Wang, Jiang, Ding, and Yang]{zhou2026touchanything}
Jianyi Zhou, Ziteng Gao, Feiyang Hong, Zirui Liu, Guannan Zhang, Weisheng Dai,
  Ruichen Zhen, Chuqiao Lyu, Haotian Wu, Yinian Mao, Xushi Wang, Yuxiang Jiang,
  Wenbo Ding, and Shuo Yang.
\newblock {TouchAnything}: A dataset and framework for bimanual tactile
  estimation from egocentric video.
\newblock arXiv preprint arXiv:2605.13083, 2026{\natexlab{a}}.
\newblock URL \url{https://arxiv.org/abs/2605.13083}.

\bibitem[Zhou et~al.(2026{\natexlab{b}})Zhou, Hong, Li, Zhao, Cen, Liu, Huang,
  Chen, Zhang, Zhu, Song, and Yang]{zhou2026touchworld}
Jianyi Zhou, Feiyang Hong, Yunhao Li, Yicheng Zhao, Yongjue Cen, Zirui Liu,
  Jiakang Huang, Zirui Chen, Ruiyang Zhang, Weizhuo Zhu, Xuhua Song, and Shuo
  Yang.
\newblock {TouchWorld}: A predictive and reactive tactile foundation model for
  dexterous manipulation.
\newblock arXiv preprint arXiv:2607.07287, 2026{\natexlab{b}}.

\end{thebibliography}
\bibliographystyle{iclr2027_conference}

\clearpage
\appendix
\section{Dataset and preprocessing}
\label{app:data}

\paragraph{Corpus.} All experiments use the \corpus{} corpus \citep{zhou2026touchanything}, released as
the HuggingFace dataset \texttt{zhouzhoujy/EgoTouch}. It records everyday bimanual manipulation with a
head-mounted wide-angle RGB camera and a pressure-sensing glove on each hand. Our copy contains 1933
episode directories from five scenes (Home, Office, Outdoor, Retail and Workbench). Each episode provides
frame-aligned egocentric RGB video at $480\times640$ and 30\,fps, two wrist-mounted RGB streams, three
sets of 3D hand-joint annotations (WiLoR, HaMeR and Rokoko), per-hand $21\times21$ pressure grids, raw
256-dimensional glove vectors, hand segmentation masks, wrist tracker poses and an episode-level contact
label. We use three of these streams: the egocentric video as context and prediction target, the
pressure grids as tactile input, and the WiLoR joints \citep{potamias2025wilor}, which are projected into
the image to render the hand-conditioning stream and to place the tactile residual.

\paragraph{Clips.} The clip planner retains 1926 of the 1933 episodes. Three episodes are shorter than
one clip and four contain video files that cannot be decoded. The retained episodes contain 2{,}183{,}052
frames, or 20.21 hours at 30\,fps. Each egocentric stream is divided into contiguous, non-overlapping
windows of 49 frames, yielding 43{,}604 clips of 1.63 seconds. Clips are stored at $432\times320$
together with a per-clip pressure tensor of shape $(49, 2, 21, 21)$ and the on-hand mask. The loader
resizes all frames to a single $384\times512$ resolution, and inference applies the same resizing, so the
evaluation geometry matches training. The stored container rate of 15\,fps is metadata only; the frames
are consecutive 30\,fps samples, so the $36$ predicted frames span $1.2$ seconds.

\paragraph{Splits.} We define our own splits rather than using the split file released with the corpus.
The held-out-episode split is disjoint from training at the episode level, and the training-episode split is disjoint
at the clip level. Because the corpus often contains several consecutive episodes of the same task, we
additionally apply a session holdout: every episode that shares a recording session with an evaluation
episode, meaning the same scene and task with consecutive timestamps at most 300 seconds apart, is removed
from training. This removes 588 episodes and 4{,}704 clips. The held-out-episode split consists of the first
clip of each of \FarmVisionOnlyOodNclips{} held-out episodes from the Home scene, covering 45 task types.
The training-episode split consists of at most the last two clips of each
of 100 long training episodes, \FarmVisionOnlyIndomNclips{} clips in total, with one 49-frame clip
discarded between the training portion of each episode and its evaluation clips so that no evaluation
clip is adjacent to a training clip. The large held-out set contains \FarmSubMethodSubNclips{} of the
4{,}704 clips removed by the session holdout, none of which is used in training; its division by task is
described below. The training split
consists of the remaining 38{,}600 clips from 1338 episodes and 174 task types, amounting to 17.51 hours
of video, with a scene composition of 53.3\% Home, 16.1\% Retail, 13.4\% Workbench, 12.8\% Outdoor and
4.4\% Office.

\paragraph{In-distribution and out-of-distribution tasks.} The session holdout makes the held-out set
disjoint from training in episodes and sessions but not in content: 39 of its 45 task names never occur in
training, yet most of these tasks have a close counterpart among the 174 training tasks. Because task instructions
are generated from the task name, we compared every held-out task name with the training task names and
assigned it to one of four groups (\Cref{tab:tasktiers}). A task is \emph{in-distribution} if the same
task, a near-duplicate or a task with a related object or action occurs in training, and
\emph{out-of-distribution} if nothing similar does. The out-of-distribution group consists of three tasks,
pushing a cart, dragging a chair and using a thermometer, with $272$ of the \FarmSubMethodSubNclips{}
clips. The held-out-episode split contains three out-of-distribution clips and the training-episode split
none, so both smaller splits are in-distribution.

\begin{table}[h]
\centering
\caption{\textbf{Only three held-out tasks have no counterpart in training.} Held-out tasks grouped by their
closest counterpart in training, by task name.}
\label{tab:tasktiers}
\footnotesize
\setlength{\tabcolsep}{4pt}
\renewcommand{\arraystretch}{1.12}
\begin{tabular}{>{\raggedright\arraybackslash}p{3.0cm}>{\raggedright\arraybackslash}p{6.6cm}cc}
\toprule
Group & Examples (held-out task: training task) & Tasks & Clips \\
\midrule
Same task & pick up toothpaste; organize a suitcase; use a microwave & 6 & 479 \\
Near-duplicate & pick up earphones: over-ear headphones; grasp sunscreen: grasp body lotion; fold shorts: fold clothes; organize medicine: sort medicine & 18 & 1{,}251 \\
Related object or action & squeeze a duck toy: pick up a toy racket; use a G-clamp: rotate a table clamp; pick up a thermos cup: wash a cup & 18 & 998 \\
\midrule
Nothing similar & push a cart; drag a chair; use a thermometer & 3 & 272 \\
\bottomrule
\end{tabular}
\end{table}

\paragraph{Task instruction.} Every clip is paired with a task instruction generated from a fixed template, ``egocentric
first-person view of two hands \{task\}, hand-object manipulation'', in which underscores in the task name
are replaced by spaces. Instructions contain between 9 and 15 words, with a mean of 11.2. The instruction
is always provided, and no model drops it during training.

\paragraph{Handling of data irregularities.} The $21\times21$ pressure grid is a hand-shaped raster
rather than the sensor topology. Of its 441 cells, 217 lie on the hand and carry finite values, and the
remaining 224 are undefined by design. The loader sets the undefined cells to zero, retains their pattern
as a fixed per-hand mask, and restricts all tactile computations to the on-hand cells. The released grids
are baseline-corrected and normalized to $[0,1]$ per episode, with separate maxima for the pressure and
flexion channels.
Hand detections are missing on 10.65\% of frames for the left hand and on 12.14\% for the right. A joint
record that is null, irregular, non-finite or entirely zero is marked invalid. Gaps of at most 30 frames
are filled by linear interpolation in 3D, whereas longer gaps leave the hand undrawn. A hand's image
location is recovered from the rendered skeleton by color thresholding and is considered present only
when at least eight pixels of its color are drawn. When a hand is absent, its tactile residual is zero and
the hand-weighted loss has no peak at that hand.

\section{Tactile sensor layout}
\label{app:sensor}

\paragraph{The $21\times21$ grid.} Tactile input is provided per frame as a pair of $21\times21$ grids,
one for each hand, and stored per clip as a float16 array of shape $(49,2,21,21)$. The grid is a
hand-shaped visualization raster and not the physical sensor layout. Exactly $217$ of its $441$ cells are
finite and lie on the hand; the other $224$ are undefined, and we set them to zero and keep their pattern
as a fixed mask. The released data are baseline-corrected and rescaled to $[0,1]$ per episode
(\Cref{app:data}).

\paragraph{Duplicate cells, pressure channels and flexion channels.} The on-hand cells are not independent
measurements. Each cell maps to one index of the glove's raw $256$-dimensional reading, and up to five
cells can share an index. Removing duplicates leaves $138$ channels for the left hand and $147$ for the
right, ordered by raw index and represented by the first cell that uses each index in row-major order.
Twelve channels per hand are shared by five cells in rows $7$ to $11$ and are finger-flexion sensors; on a
random sample of $800$ training clips such a channel exceeds $0.1$ on $27.7\%$ of channel-frames, compared
with $9.9\%$ for a pressure taxel. Three further channels per hand also behave as flexion sensors although
they are shared by only two cells. They are located on the thumb block in rows $11$ and $12$, and they are
channels $93$ to $95$ on the left hand and $42$ to $44$ on the right. Over $60$ episodes they exceed $0.15$
while every other cell of the thumb block remains below $0.02$ on $0.3\%$ of frames, compared with $0.4\%$
for the five-cell flexion channels and $0.0\%$ for any pressure taxel, and they exceed $0.1$ on $26.5\%$ of
channel-frames, compared with $12.7\%$ for a fingertip taxel. Each hand therefore has $15$ flexion
channels, three per digit, together with $123$ pressure taxels on the left and $132$ on the right. All
channels are provided to the model; the distinction only affects how the readings are displayed in the
figures and how contact is measured in \Cref{app:contact}.

\paragraph{Standardization.} Each channel is standardized with a mean and standard deviation computed
over training frames in the $[0,1]$ grid space. The standard deviation is bounded below by $10^{-3}$, and
the standardized value is clipped to $[-8, 8]$. One left-hand channel and eighteen right-hand channels
are constant in the training data and remain at the lower bound. These statistics were computed on
$1{,}282$ clips of the Home scene. The per-frame embeddings, rather than the channel values, are averaged within
each temporal group of the video autoencoder, which reduces $49$ pixel frames to $13$ latent frames.

\paragraph{Validation of region positions.} Region positions are not read from the raster. Fingertips
are joints $4$, $8$, $12$, $16$ and $20$, and the palm is the mean of joints $0$, $5$, $9$, $13$ and $17$,
projected with the same function that renders the skeleton, so the two are aligned by construction. A
region is invalid when the hand is undetected or its projected point lies outside the image. The
resulting array of shape $(49,2,6,3)$ was computed for all $1926$ episodes without failure. At the start
of training we verified that the region path was active, that the validity of each region was determined
independently, and that the injected residual was exactly zero.

\section{Model and training details}
\label{app:training}

\Cref{tab:hparams} lists the complete training recipe. \method{}, the capacity-matched control, every
ablation and every other visual-tactile world model use these settings and differ only in the tactile pathway under
study.

\paragraph{Backbone and geometry.} \method{} finetunes Wan2.2-Fun-5B-InP \citep{wan2025video}, a
flow-matching video transformer with $5{,}001{,}016{,}512$ parameters, hidden dimension 3072, 30 blocks,
24 attention heads of dimension 128 and a feed-forward width of 14{,}336. A training example is a clip of
49 frames at $384\times512$, which the causal video autoencoder compresses by a factor of four in time and
sixteen in space into 48-channel latents; a non-overlapping $(1,2,2)$ patchifier turns these into $2496$
video tokens. Apart from text cross-attention and the tactile residual, all conditioning is by channel
concatenation. The pretrained inpainting input layer reads 100 channels, namely 48 noisy latents, a
4-channel observation mask and 48 reference latents. We extend the patch embedding to 148 channels by
appending the encoded hand-skeleton stream, copy the pretrained weights into the first 100 input channels
and initialize the 48 new channels to zero, so the extended model is functionally identical to the
pretrained model at initialization. Text conditioning uses a frozen umt5-xxl encoder with a 512-token
context. The noise schedule is a flow-matching Euler discrete scheduler with 1000 training timesteps and
shift 12.0, and the same scheduler is used at inference (\Cref{app:protocol}).

\paragraph{Trained parameters.} The autoencoder, the text encoder and the transformer weights are
frozen. Adaptation uses two disjoint sets of parameters. The first is a LoRA of rank 64 and scale 64
\citep{hu2022lora} on all 306 linear modules of the transformer. The second consists of densely trained
modules: the extended patch embedding, the tactile embedding and the five-block spatial injector. Modules
of the tactile pathway are excluded from the LoRA, so the pathway is trained at full rank and does not
share LoRA capacity. The capacity-matched control instantiates and trains exactly the same modules with
the same initialization. Its pressure tensor is set to zero before the tactile embedding at every training
and evaluation step. The embedding still adds its learned frame and hand terms, so the control's residual
is a trainable signal placed at the hands that carries no sensor information. The two models therefore
differ only in whether the reading contains the sensor's signal.

\paragraph{Optimizer, precision and schedule.} A single AdamW optimizer \citep{loshchilov2019decoupled}
updates both parameter sets with a learning rate of $3\times10^{-5}$, $\beta = (0.9, 0.999)$,
$\epsilon = 10^{-8}$ and weight decay $0.01$. The learning rate is constant after a linear warmup of 100
steps, and gradients are clipped to a global norm of $1.0$. Training uses bf16 precision with gradient
checkpointing.

\paragraph{Distributed setup.} Each run uses two GPUs with \texttt{accelerate} and DeepSpeed ZeRO-2
\citep{rajbhandari2020zero}, with optimizer states offloaded to the CPU. Each GPU processes one clip per
step and the parameters are updated after every step, so the effective batch size is two clips and one
epoch comprises $19{,}300$ updates. Peer-to-peer and InfiniBand transports are disabled, and all-reduce
operations pass through host memory. All models use identical settings.

\paragraph{Budget and hardware.} Each model is finetuned for exactly one epoch over the $38{,}600$ training
clips, and the checkpoint at the end of the epoch is evaluated. Validation during training is disabled, so
all reported evaluations come from a separate inference pass. Each run uses two NVIDIA RTX 6000 Ada GPUs
with 49\,GB of memory, of which about 38\,GB per GPU are used, and takes between 5\,h\,20 and 6\,h\,30.

\begin{table}[t]
\centering
\caption{\textbf{One recipe for every model in the paper.} Complete training recipe.}
\label{tab:hparams}
\small
\begin{tabular}{@{}lp{0.60\textwidth}@{}}
\toprule
\multicolumn{2}{@{}l}{\textit{Backbone}} \\
Pretrained model & Wan2.2-Fun-5B-InP (flow-matching video transformer) \\
Transformer parameters & $5{,}001{,}016{,}512$ (after the channel extension) \\
Hidden dimension, blocks, heads & 3072, 30, 24 (head dimension 128) \\
Feed-forward dimension & 14{,}336 \\
Patchifier & 3D convolution, kernel and stride $(1,2,2)$ \\
Video autoencoder & 48 latent channels, $4\times$ temporal and $16\times$ spatial compression \\
Clip length and resolution & 49 frames, $384\times512$ \\
Latent and token grid & $13\times24\times32$ latents, $13\times12\times16 = 2496$ tokens \\
Patch-embedding channels & $148$: 48 noise, 4 mask, 48 reference, 48 hand skeleton \\
Text encoder & umt5-xxl, frozen, 512 tokens \\
Noise schedule & flow-matching Euler discrete, 1000 training timesteps, shift 12.0 \\
\midrule
\multicolumn{2}{@{}l}{\textit{Trained parameters}} \\
LoRA rank, scale, targets & 64, 64, all 306 linear modules of the transformer \\
LoRA parameters & $164{,}261{,}888$ \\
Dense parameters & patch embedding $1{,}821{,}696$; tactile embedding $4{,}349{,}725$; injector $47{,}232{,}000$ \\
Total trainable parameters & $217{,}665{,}309$ ($4.35\%$), identical for \method{}, the control and every tactile input \\
\midrule
\multicolumn{2}{@{}l}{\textit{Optimization}} \\
Optimizer & AdamW, $\beta = (0.9, 0.999)$, $\epsilon = 10^{-8}$, weight decay $0.01$ \\
Learning rate & $3\times10^{-5}$, constant after a 100-step warmup \\
Gradient clipping & global norm $1.0$ \\
Precision & bf16, gradient checkpointing \\
Clips per GPU and number of GPUs & 1 and 2 \\
Effective batch size & 2 clips per update \\
Updates per epoch & $19{,}300$ \\
Budget & 1 epoch, $38{,}600$ clips \\
\midrule
\multicolumn{2}{@{}l}{\textit{Systems}} \\
Software & \texttt{accelerate} 1.2.1, DeepSpeed 0.19.2 (ZeRO-2, CPU optimizer offload), PyTorch 2.6.0 \\
Hardware per run & 2 NVIDIA RTX 6000 Ada (49\,GB), about 38\,GB used per GPU \\
Duration per run & 5\,h\,20 to 6\,h\,30 \\
\bottomrule
\end{tabular}
\end{table}

\subsection{Further details}
\label{app:training:deferred}
\label{app:zeroinit}

\paragraph{Alignment of latent and pixel frames.} The causal autoencoder maps the 49 pixel frames of a
clip to 13 latent frames: latent frame $0$ contains pixel frame $0$, and latent frame $k>0$ contains pixel
frames $4k-3$ to $4k$. This grouping determines the boundary of the observation, since pixel frames
$0$ to $K$ correspond to latent frames $0$ to $m=K/4$, and it determines the temporal reduction of the
tactile embeddings (\Cref{app:sensor}).

\paragraph{Why a gated projection does not train.} An earlier version of the injector used a randomly
initialized projection behind a per-block scalar gate, setting the scale in \Cref{eq:residual} to
$\lambda^{(\ell)} = \tanh(\gamma^{(\ell)})$ with $\gamma^{(\ell)}$ initialized to zero. The gradient with
respect to $W^{(\ell)}$ then contains the factor $\tanh(\gamma^{(\ell)})$, which is zero at
initialization, so the projection receives no gradient and retains its random values. The gate can move,
but opening it only injects a random projection of the reading, so nothing drives it open. We confirmed
numerically that the projection's gradient was zero and that the gates remained closed. Initializing the
projection itself to zero, as in zero convolutions \citep{zhang2023controlnet} and zero-initialized
low-rank adapters \citep{hu2022lora}, keeps the residual exactly zero at initialization while providing
gradient to the projection from the first step.

\paragraph{Loss weighting.} Over latent frames, the loss weight is $0$ on observed latents $0,\dots,m$,
$1$ on the following chunk $m+1,\dots,m+C$ and $0.2$ on all later latents. Spatially, the weight is
multiplied by $1 + 5.0\,\mathcal{H}$, where $\mathcal{H}\in[0,1]$ is the maximum over both hands of a
unit-peak Gaussian of width $2.4$ latent cells centered at the hand's latent location. This weighting uses
ground-truth hand locations of the predicted frames only as supervision, and it is never provided as input.

\subsection{Training schedules and variants}
\label{app:recipes}

\paragraph{Three-run comparison.} The full-reading model and the control were each trained three
times with the recipe of \Cref{tab:hparams}, and \Cref{tab:perrun:app} lists the result of each run.

\paragraph{Tactile inputs.} The inputs of \Cref{tab:input} were trained in a later round with the recipe
of \Cref{tab:hparams}, one run is reported for each input, and all were evaluated on the session-held-out set. They differ from
\method{} only in the values written into the $21\times21$ grids before the embedding. \emph{Per-taxel
contact} thresholds each pressure taxel at $0.05$ and writes a constant, leaving the flexion channels
unchanged, so only the force magnitude is removed. The \emph{contact bit} sets every cell of a hand to that
constant when more than $2\%$ of the hand's pressure taxels exceed $0.05$ and to zero otherwise.

\subsection{Other visual-tactile world models}
\label{app:reimpl}

We compare the ways in which existing visual-tactile world models feed a tactile signal to a video
predictor, not complete systems. The three
original systems differ from ours in predictor, embodiment and task (\Cref{tab:reimpl}), so their reported
numbers are not comparable with ours. TouchWorld pretrains on the same corpus as ours; VT-WM and
FeelWorld use other sensors on a robot hand or gripper. For each we keep the way the tactile signal is fed to the
predictor and take everything else from \method{}: the pretrained backbone and its LoRA adapter, the
hand-skeleton conditioning, the training clips, the recipe of \Cref{tab:hparams}, the sampling of the
observation length, the restriction of the tactile input to frames $0$ to $K$, and the evaluation. All three
receive the glove reading of \method{}, standardized as in \Cref{sec:method:injection}, which is the
signal TouchWorld was pretrained on and replaces the original sensor of VT-WM and FeelWorld, so the rows
of \Cref{tab:compare} differ only in how the tactile signal is fed to the model. Each was trained once.

\begin{table}[h]
\centering
\caption{\textbf{Each existing visual-tactile world model keeps its original mechanism and receives the
same glove reading as \method{}.} The original system and our implementation.}
\label{tab:reimpl}
\footnotesize
\setlength{\tabcolsep}{4pt}
\renewcommand{\arraystretch}{1.15}
\begin{tabular}{>{\raggedright\arraybackslash}p{2.15cm}>{\raggedright\arraybackslash}p{3.5cm}>{\raggedright\arraybackslash}p{3.5cm}>{\raggedright\arraybackslash}p{3.5cm}}
\toprule
& TouchWorld \citep{zhou2026touchworld} & VT-WM \citep{higuera2026visuotactile} & FeelWorld \citep{ma2026feelworld} \\
\midrule
Original sensor & the same pressure gloves (our corpus), and tactile gloves on two robot hands & four Digit~360 optical fingertip sensors on an Allegro hand & DM tactile sensors, 3D contact point cloud \\
Original predictor & video model finetuned from Wan2.2-TI2V-5B & action-conditioned latent transformer over Cosmos tokens & latent dynamics predictor over V-JEPA~2 features \\
Original mechanism & readings rendered as tactile images beside the RGB views & Sparsh-X tactile tokens concatenated with the visual tokens & one global tactile token per fingertip sensor, attended by the visual tokens through a contact-gated attention \\
\midrule
Our implementation & grids drawn as image panels in the hand-skeleton conditioning stream & one token per latent frame and hand appended to the video tokens & one tactile feature per hand added to all video tokens, scaled by a learned gate \\
Tactile parameters & $51{,}581{,}725$ (zeroed residual pathway kept) & $23{,}236{,}381$ & $51{,}597{,}090$ \\
\bottomrule
\end{tabular}
\end{table}

\paragraph{TouchWorld: rendered panels.} Each hand's $21\times21$ grid, including the flexion cells, is
clipped to $[0,1]$ and drawn in grayscale at $4$ pixels per cell, an $84\times84$ panel with off-hand
cells black and a thin gray frame, $6$ pixels from the top-left corner (left hand) and the top-right corner
(right hand) of the hand-skeleton video. Grayscale keeps the panels out of the red and green that identify
the two hands in the skeleton render. The panels therefore reach the transformer through the same widened
patch embedding as the skeleton, at a fixed image position rather than at the hand, and are blank after
frame $K$ together with the skeleton. The residual pathway of \Cref{sec:method:injection} is kept with a
zeroed reading, exactly as in the control, so this model is the control with the panels added.

\paragraph{VT-WM: appended tokens.} The standardized readings are averaged over the four pixel frames of
each latent frame and embedded by the per-hand linear map of \Cref{sec:method:injection} together with the
frame, hand and observed-frame embeddings. A residual two-layer network (LayerNorm, linear, SiLU, linear,
with a zero-initialized output layer) stands in for the original tactile encoder. This gives one token
per latent frame and hand, $26$ per clip, which are appended after the $2{,}496$ video tokens and processed
by all thirty blocks together with them. Readings after the observation are set to the standardized
mean, and the block-causal mask is extended with the chunk index of each tactile token, so that a video
token attends to the tactile tokens of its own and earlier chunks only. The tactile tokens are discarded at
the output. There is no per-block projection, which is why this design has fewer tactile parameters.

\paragraph{FeelWorld: gated global feature.} The tactile embedding is projected at the five injection
blocks as in \Cref{eq:residual}, but the footprint $\hat g$ is replaced by a uniform distribution over the
$12\times16$ token grid, so each hand contributes one global feature per latent frame. At each injection
block the projected feature is multiplied by a gate $\mathrm{sigmoid}(w^{(\ell)\top}\mathrm{LN}^{(\ell)}(e)
+ c^{(\ell)})$ predicted from the tactile embedding, which adds $5\times3{,}073 = 15{,}365$ parameters. The
original model gates the attention from the visual tokens to its tactile tokens by a separately
predicted contact probability; we keep the residual form shared by the other models and learn the gate end
to end, so that this model differs from them only in the gate and in the absence of placement.

\section{Evaluation protocol}
\label{app:protocol}

\paragraph{Deployment setting and sampling.} All generated-video results in this paper use one evaluation
setting. A 49-frame clip is divided at $K=12$: pixel frames $0$ to $12$, that is, thirteen frames spanning
$0.43$ seconds and four latent frames, form the observation, and the model denoises the remaining nine
latent frames, corresponding to 36 pixel frames, in a single pass. Observed latents are
held at their clean encodings with timestep zero throughout sampling. Attention is block-causal over chunks
of four latent frames, so the first predicted chunk (pixel frames $13$ to $28$) attends only to observed
latents, the second (frames $29$ to $44$) attends in addition to the first, and the third (frames $45$ to
$48$) attends to everything before it. The hand skeleton is black and the tactile residual is zero after
frame $K$, so no future information is supplied. Sampling uses 30 flow-matching steps with scheduler shift
$12.0$ and classifier-free guidance $6.0$ against the negative prompt ``bad detailed''. The footprint
width at evaluation is the inference default $\sigma=3.0$, whereas training uses $\sigma=1.8$; evaluating
one checkpoint at both widths changes LPIPS by
\DconSigmaWidthOodLpips{} on the held-out-episode split, and all models share the
same evaluation width. Generated frames at $384\times512$ are resized bilinearly to the stored
$320\times432$ resolution before scoring. All 36 predicted frames are scored. The analysis in
\Cref{sec:exp:horizon} scores shorter windows of the same generated videos without regenerating them.

\paragraph{Metrics.} LPIPS uses the AlexNet backbone with the calibrated linear head, inputs mapped to
$[-1,1]$, computed per frame and averaged over the scored frames. LPIPS$_{\text{hand}}$ applies the same
metric to one crop per clip, defined as the union of rendered skeleton pixels over the scored window,
enlarged by $25\%$ of its height and width on each side and resized to $128\times128$. Because the enlarged
crop usually contains the grasped object, it measures the manipulation region rather than the hands
alone, and it is undefined for the few clips without a usable skeleton. PSNR is computed as
$10\log_{10}(255^{2}/\mathrm{MSE})$ over full RGB frames. SSIM is the scikit-image implementation on the
mean of the color channels with a data range of 255 and default settings.

\paragraph{Statistical procedure.} The unit of pairing is the clip. For a model with several runs, the
per-clip scores are first averaged over its runs and then compared with the other model, so that
run-to-run variation is included in the estimate rather than being treated as signal. The vector of per-clip
differences, with fewer entries where LPIPS$_{\text{hand}}$ is undefined, is resampled with replacement
$20{,}000$ times with a fixed generator. We report the mean difference and the $2.5$ and $97.5$
percentiles of the resampled means. For
sets of more than $400$ clips the normal approximation to this interval is used, which agrees with the
bootstrap to the reported precision. The interval accounts for clip sampling and evaluation noise but not
for the variation between independently trained runs, which is therefore reported separately through
the results of individual runs (\Cref{tab:perrun:app}).

\subsection{Hand-behavior metrics}
\label{app:handmetrics}

Frame-level perceptual metrics are dominated by background texture and are largely insensitive to whether
the hands behave correctly. We therefore measure hand behavior directly, computing every quantity with the
same estimator on ground-truth and generated video so that estimator bias cancels in the difference. The
glove silhouette is extracted as dark pixels inside a crop centered on the ground-truth hand location. The
crop is necessary because, over the full frame, the wearer's body and shadows dominate the dark pixels.
From the silhouette we compute its overlap with the ground truth (IoU), the distance between centroids in
pixels, the ratio of generated to ground-truth silhouette area (ideal value one), and the dynamic time
warping distance between the two centroid trajectories (DTW). Optical flow within the same crop gives the
agreement of flow directions and the ratio of total flow magnitude to the ground truth
(ideal value one). Frames are averaged within each clip before pooling, since pooling individual frames
would treat correlated observations as independent. The window is the full predicted interval, frames $13$
to $48$.

\subsection{Sharpness}
\label{app:sharpness}

For \Cref{sec:exp:looks} we compute, for each frame, the variance of the Laplacian of the grayscale image,
a standard measure of focus, and for each clip the ratio of the mean over generated frames to the mean over
ground-truth frames within the scored window. On the \SharpMethodN{}-clip set, this ratio is
\SharpControlRatio{} for the capacity-matched control, \SharpMethodRatio{} for the full-reading model,
and \SharpGateZeroRatio{} for that checkpoint with its tactile pathway disabled
(\SharpMethodVsControlDelta{}\,\SharpMethodVsControlCi{} for \method{} relative
to the control). No model exceeds the
sharpness of the ground truth, so the ordering reflects how much of the ground-truth detail each model
reproduces rather than the addition of spurious texture.

\section{Complete results}
\label{app:results}

This appendix contains the complete tables for \Cref{sec:exp}. Unless stated otherwise, all scores are
computed over the 36 predicted frames of the 49-frame clips. LPIPS and LPIPS$_{\text{hand}}$ are better
when lower, and PSNR and SSIM when higher. Every difference is clip-paired
(\Cref{app:protocol}). Models are comparable within a table; absolute scores
from different training rounds or clip lengths should not be compared.

\begin{table}[h]
\centering
\caption{\textbf{Three independent training runs per model.} Mean $\pm$ standard
deviation over runs. The difference is clip-paired.}
\label{tab:main}
\footnotesize
\setlength{\tabcolsep}{4pt}
\begin{tabular}{lcccc}
\toprule
& \multicolumn{2}{c}{Held-out episodes} & \multicolumn{2}{c}{Training episodes} \\
\cmidrule(lr){2-3}\cmidrule(lr){4-5}
Model & LPIPS & LPIPS$_{\text{hand}}$ & LPIPS & LPIPS$_{\text{hand}}$ \\
\midrule
Vision-only baseline & 0.4905\,$\pm$\,0.0328 & 0.3817\,$\pm$\,0.0439 & 0.5024\,$\pm$\,0.0417 & 0.3992\,$\pm$\,0.0488 \\
\method{} (ours) & \textbf{0.4654}\,$\pm$\,0.0109 & \textbf{0.3536}\,$\pm$\,0.0119 & \textbf{0.4777}\,$\pm$\,0.0170 & \textbf{0.3733}\,$\pm$\,0.0149 \\
\midrule
Difference & $-0.0252$ & $-0.0281$ & $-0.0247$ & $-0.0259$ \\
\bottomrule
\end{tabular}

\end{table}

\paragraph{Individual runs.} \Cref{tab:perrun:app} lists the mean LPIPS of each run of the full-reading
model and of the control in the three-run comparison. Because the runs of the two models are trained
independently, we compare every run of \method{} with every run of the control. The pooled clip-paired
interval of \Cref{tab:main} accounts for clip sampling and evaluation noise; this table shows the variation
between training runs.

\begin{table}[h]
\centering
\caption{\textbf{\method{} wins 7 of the 9 run pairings on each split, and its worst run is below the
baseline's mean.} Mean LPIPS of each independently trained run in the main comparison, and the number of
run comparisons in which \method{} obtains the lower score.}
\label{tab:perrun:app}
\footnotesize
\setlength{\tabcolsep}{5pt}
\begin{tabular}{lcccc}
\toprule
& \multicolumn{2}{c}{Held-out episodes} & \multicolumn{2}{c}{Training episodes} \\
\cmidrule(lr){2-3}\cmidrule(lr){4-5}
& control & \method{} & control & \method{} \\
\midrule
run 1 & 0.5281 & 0.4777 & 0.5505 & 0.4963 \\
run 2 & 0.4677 & 0.4571 & 0.4771 & 0.4630 \\
run 3 & 0.4758 & 0.4612 & 0.4796 & 0.4737 \\
\midrule
mean over runs & 0.4905 & 0.4654 & 0.5024 & 0.4777 \\
run comparisons won by \method{} & \multicolumn{2}{c}{7 of 9} & \multicolumn{2}{c}{7 of 9} \\
\bottomrule
\end{tabular}

\end{table}

\paragraph{Scoring window.} \Cref{tab:curve:app} contains the values shown in \Cref{fig:horizon}. The
upper block scores the same predictions over the first 8, 16, 24 and 36 predicted frames. The lower block
gives the mean per frame within each block of frames, obtained by differencing cumulative means:
$\bar\ell_{a+1..b} = (b\,\bar\ell_{1..b} - a\,\bar\ell_{1..a})/(b-a)$, with indices counted over predicted
frames.

\begin{table}[h]
\centering
\caption{\textbf{The per-frame improvement is smallest in the first block and several times larger in the
later blocks.} LPIPS as a function of the scored window, and per frame
within each block of predicted frames, averaged over the runs of \Cref{tab:main}.}
\label{tab:curve:app}
\footnotesize
\setlength{\tabcolsep}{3.6pt}
\resizebox{\textwidth}{!}{\begin{tabular}{lcccccc}
\toprule
& \multicolumn{3}{c}{Held-out episodes} & \multicolumn{3}{c}{Training episodes} \\
\cmidrule(lr){2-4}\cmidrule(lr){5-7}
Predicted frames scored & control & \method{} & difference & control & \method{} & difference \\
\midrule
\multicolumn{7}{l}{\emph{Cumulative window, mean over the frames scored}} \\
frames 13 to 20 & 0.2727 & \textbf{0.2644} & $-0.0083$ & 0.2812 & \textbf{0.2770} & $-0.0042$ \\
frames 13 to 28 & 0.3593 & \textbf{0.3501} & $-0.0092$ & 0.3624 & \textbf{0.3559} & $-0.0065$ \\
frames 13 to 36 & 0.4265 & \textbf{0.4087} & $-0.0178$ & 0.4329 & \textbf{0.4168} & $-0.0161$ \\
frames 13 to 48 & 0.4905 & \textbf{0.4654} & $-0.0252$ & 0.5024 & \textbf{0.4777} & $-0.0247$ \\
\midrule
\multicolumn{7}{l}{\emph{Mean per frame within each block}} \\
frames 13 to 20 (first chunk) & 0.2727 & \textbf{0.2644} & $-0.0083$ & 0.2812 & \textbf{0.2770} & $-0.0042$ \\
frames 21 to 28 (first chunk) & 0.4459 & \textbf{0.4358} & $-0.0101$ & 0.4436 & \textbf{0.4348} & $-0.0088$ \\
frames 29 to 36 (second chunk) & 0.5610 & \textbf{0.5260} & $-0.0350$ & 0.5739 & \textbf{0.5386} & $-0.0353$ \\
frames 37 to 48 (second and third chunks) & 0.6185 & \textbf{0.5786} & $-0.0399$ & 0.6414 & \textbf{0.5994} & $-0.0420$ \\
\bottomrule
\end{tabular}
}
\end{table}

\paragraph{Hand behavior.} \Cref{tab:hand:app} reports the hand-behavior metrics of
\Cref{app:handmetrics} over the full predicted window, for \method{} and the vision-only baseline on the
same \FarmSubMethodSubNclips{} clips.

\paragraph{Metrics of \Cref{fig:teaser}.} \Cref{tab:radar:app} lists the six metrics shown in the
radar chart of \Cref{fig:teaser} for every model on the held-out set. Each value is averaged over clips.

\begin{table}[h]
\centering
\caption{\textbf{\method{} is ahead of every existing visual-tactile world model on all six metrics.} The metrics of
\Cref{fig:teaser} on the \FarmSubMethodSubNclips{}-clip held-out set, predicted frames 13 to 48.}
\label{tab:radar:app}
\footnotesize
\setlength{\tabcolsep}{4pt}
\begin{tabular}{lcccccc}
\toprule
Model & LPIPS $\downarrow$ & LPIPS$_{\text{hand}}$ $\downarrow$ & IoU $\uparrow$ & centroid (px) $\downarrow$ & DTW (px) $\downarrow$ & flow dir. $\uparrow$ \\
\midrule
TouchWorld & 0.494 & 0.400 & 0.376 & 28.0 & 14.2 & 0.031 \\
FeelWorld & 0.454 & 0.351 & 0.430 & 25.1 & 11.9 & 0.027 \\
VT-WM & 0.429 & 0.337 & 0.462 & 23.3 & 11.2 & 0.032 \\
\midrule
\method{} (ours) & \textbf{0.421} & \textbf{0.327} & \textbf{0.476} & \textbf{21.9} & \textbf{10.4} & \textbf{0.041} \\
\bottomrule
\end{tabular}

\end{table}

\begin{table}[h]
\centering
\caption{\textbf{Touch changes how much of the hands is drawn and how much they move: the silhouette area
and flow magnitude ratios move toward one.} Hand-behavior metrics over frames 13 to 48 on the held-out set,
clip-paired.}
\label{tab:hand:app}
\footnotesize
\setlength{\tabcolsep}{3.0pt}
\begin{tabular}{lccc}
\toprule
& \multicolumn{3}{c}{\method{} against the vision-only baseline} \\
\cmidrule(lr){2-4}
Metric & vision-only & \method{} & difference \\
\midrule
silhouette IoU $\uparrow$ & 0.475 & \textbf{0.476} & $+0.001$ \\
silhouette area ratio (ideal 1) & 0.920 & \textbf{1.032} & $+0.111$ \\
trajectory DTW (px) $\downarrow$ & 10.41 & \textbf{10.37} & $-0.04$ \\
flow direction agreement $\uparrow$ & 0.039 & \textbf{0.041} & $+0.003$ \\
flow magnitude ratio (ideal 1) & 0.771 & \textbf{0.912} & $+0.141$ \\
\bottomrule
\end{tabular}

\end{table}

\subsection{PSNR and SSIM}
\label{app:pixelmetrics}

\Cref{tab:compare:app,tab:main:app} repeat the main-text tables with PSNR and SSIM added.
On every split the vision-only baseline has the higher PSNR and SSIM while \method{} has the lower LPIPS,
and the existing visual-tactile world models are ordered differently by the two kinds of metric as well. The pixel metrics are
reported here rather than in the main text because on this task they reward a prediction for committing
to less, which \Cref{tab:sanity} shows directly. Blurring the output of \method{} with a Gaussian of
increasing width raises its PSNR from \SanityDtwmPsnr{} to \SanityBlurThreePsnr{} and its SSIM from
\SanityDtwmSsim{} to \SanityBlurThreeSsim{}, while its LPIPS rises from \SanityDtwmLpips{} to
\SanityBlurThreeLpips{}; the blurred prediction scores higher on PSNR on \SanityBlurThreePsnrWin{} of the
clips and on SSIM on \SanityBlurThreeSsimWin{}.

The direction of the gap between the two models follows from the same property. Of the \ProbeRawGap{} dB
by which \method{} trails the baseline on \ProbeN{} clips, \ProbeGainShare{} disappears when each predicted
frame's mean and standard deviation are matched to the ground truth's before scoring, and only
\ProbeBlurShare{} when prediction and ground truth are both blurred: the gap is carried by global
brightness and contrast, not by the high-frequency detail measured in \Cref{app:sharpness}. On
\PhotoN{} clips the baseline's frames are lower in contrast and in saturation than the ground truth by
\PhotoBaselineStd{} and \PhotoBaselineSat{} grey levels, those of \method{} by \PhotoDtwmStd{} and
\PhotoDtwmSat{}. Under a squared error, shrinking contrast toward the mean is the optimal hedge for an
uncertain prediction, so the model whose frames are closer to the ground truth in contrast pays for it in
PSNR and SSIM.

\begin{table}[h]
\centering
\caption{\textbf{PSNR and SSIM improve when the prediction is blurred; LPIPS penalizes blur.} LPIPS, PSNR and SSIM of the two models and of degraded versions of the \method{}
prediction on \SanityN{} held-out clips, frames 13 to 48; the last row is the models' reconstruction of the
observed frames, which no prediction can pass. Best value of each column in bold.}
\label{tab:sanity}
\footnotesize
\setlength{\tabcolsep}{5pt}
\begin{tabular}{lccc}
\toprule
Prediction & LPIPS $\downarrow$ & PSNR $\uparrow$ & SSIM $\uparrow$ \\
\midrule
Vision-only baseline & 0.414 & 15.24 & 0.547 \\
\method{} & \textbf{0.412} & 14.94 & 0.535 \\
\midrule
\method{}, Gaussian blur $\sigma = 1$ & 0.476 & 15.08 & 0.551 \\
\method{}, Gaussian blur $\sigma = 2$ & 0.565 & 15.21 & 0.564 \\
\method{}, Gaussian blur $\sigma = 3$ & 0.625 & \textbf{15.30} & \textbf{0.571} \\
\midrule
reconstruction of the observed frames & 0.031 & 35.50 & 0.960 \\
\bottomrule
\end{tabular}

\end{table}

\begin{table}[h]
\centering
\caption{\textbf{PSNR and SSIM order the models differently from LPIPS.} \Cref{tab:compare} with PSNR and
SSIM added, on the same clips.}
\label{tab:compare:app}
\footnotesize
\setlength{\tabcolsep}{3.0pt}
\resizebox{\textwidth}{!}{\begin{tabular}{lccccccccc}
\toprule
& & \multicolumn{4}{c}{ID (2,728 clips)} & \multicolumn{4}{c}{OOD (272 clips)} \\
\cmidrule(lr){3-6}\cmidrule(lr){7-10}
Model & Params & LPIPS $\downarrow$ & LPIPS$_{\text{hand}}$ $\downarrow$ & PSNR $\uparrow$ & SSIM $\uparrow$ & LPIPS $\downarrow$ & LPIPS$_{\text{hand}}$ $\downarrow$ & PSNR $\uparrow$ & SSIM $\uparrow$ \\
\midrule
Vision-only baseline & 51.6M & 0.4243 & 0.3273 & \textbf{15.28} & 0.5095 & 0.4882 & 0.3765 & \textbf{13.99} & 0.5144 \\
TouchWorld \citep{zhou2026touchworld} & 51.6M & 0.4899 & 0.3979 & 14.80 & 0.5088 & 0.5341 & 0.4204 & 13.88 & \textbf{0.5218} \\
VT-WM \citep{higuera2026visuotactile} & 23.2M & 0.4237 & 0.3324 & 15.11 & \textbf{0.5102} & 0.4854 & 0.3797 & 13.91 & 0.5203 \\
FeelWorld \citep{ma2026feelworld} & 51.6M & 0.4488 & 0.3465 & 14.40 & 0.4896 & 0.5071 & 0.3964 & 13.36 & 0.5012 \\
\midrule
\method{} (ours) & 51.6M & \textbf{0.4157} & \textbf{0.3227} & 15.06 & 0.4994 & \textbf{0.4774} & \textbf{0.3668} & 13.87 & 0.5068 \\
\bottomrule
\end{tabular}
}
\end{table}

\begin{table}[h]
\centering
\caption{\textbf{Over three runs, PSNR and SSIM favor the baseline where LPIPS favors \method{}.}
\Cref{tab:main} with PSNR and SSIM added, three runs per model.}
\label{tab:main:app}
\footnotesize
\setlength{\tabcolsep}{3.0pt}
\resizebox{\textwidth}{!}{\begin{tabular}{lcccccccc}
\toprule
& \multicolumn{4}{c}{Held-out episodes} & \multicolumn{4}{c}{Training episodes} \\
\cmidrule(lr){2-5}\cmidrule(lr){6-9}
Model & LPIPS $\downarrow$ & LPIPS$_{\text{hand}}$ $\downarrow$ & PSNR $\uparrow$ & SSIM $\uparrow$ & LPIPS $\downarrow$ & LPIPS$_{\text{hand}}$ $\downarrow$ & PSNR $\uparrow$ & SSIM $\uparrow$ \\
\midrule
Vision-only baseline & 0.4905 & 0.3817 & \textbf{14.98} & \textbf{0.5271} & 0.5024 & 0.3992 & 13.83 & \textbf{0.4751} \\
\method{} (ours) & \textbf{0.4654} & \textbf{0.3536} & 14.89 & 0.5180 & \textbf{0.4777} & \textbf{0.3733} & \textbf{13.95} & 0.4705 \\
\midrule
Difference & $-0.0252$ & $-0.0281$ & $-0.09$ & $-0.0091$ & $-0.0247$ & $-0.0259$ & $+0.12$ & $-0.0045$ \\
\bottomrule
\end{tabular}
}
\end{table}

\section{Qualitative results}
\label{app:qualitative}

\begin{figure}[t]
\centering
\includegraphics[width=\textwidth]{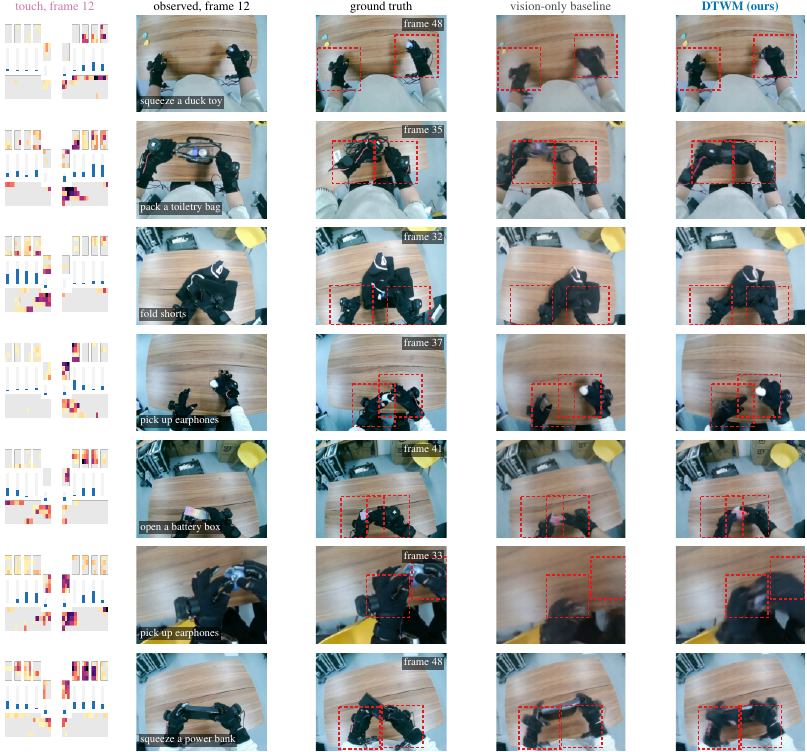}
\caption{\textbf{Further in-distribution examples show the same behavior as \Cref{fig:id}: the hands stay in
place with the reading and drift or dissolve without it.} As \Cref{fig:id}.}
\label{fig:id:app}
\end{figure}

\begin{figure}[t]
\centering
\includegraphics[width=\textwidth]{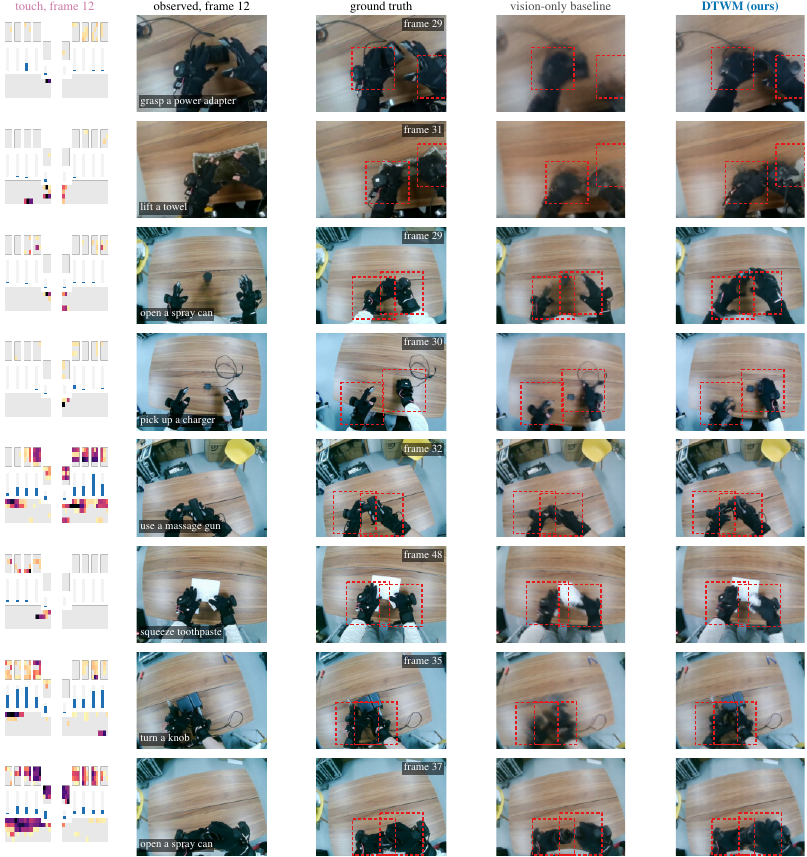}
\caption{\textbf{The same behavior on further in-distribution examples}, continued from \Cref{fig:id:app}.}
\label{fig:id:app2}
\end{figure}

\begin{figure}[t]
\centering
\includegraphics[width=\textwidth]{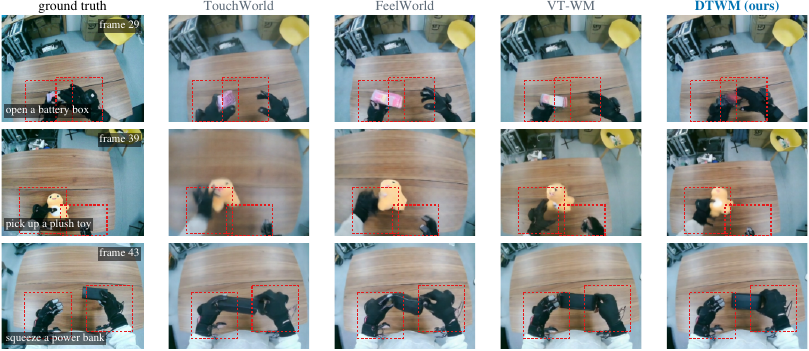}
\caption{\textbf{On further clips, the hands of \method{} remain closest to the ground truth among the
tactile models.} As \Cref{fig:methods}.}
\label{fig:methods:app}
\end{figure}

Generated frames are shown at the $384\times512$ resolution produced by the model.

\section{Contact and force dynamics of the corpus}
\label{app:contact}

This appendix measures how the tactile signal changes over the frames the model predicts, and whether its
course over the observed frames carries information about what happens next. All statistics are computed on
\ForceNClips{} training clips in the evaluation layout of \Cref{app:protocol}: the observation ends at frame
$12$, the first predicted chunk covers frames $13$ to $28$ and the second frames $29$ to $44$.

\paragraph{Definition.} Contact and force are measured on the pressure taxels of each hand, excluding the
fifteen flexion channels (\Cref{app:sensor}). A taxel is pressed when its normalized pressure exceeds
$0.05$, and a hand is in contact when more than $2\%$ of its taxels are pressed. The force of a hand is the
sum of its taxel values, the observed trend is the change of that force from frames $0$ to $3$ to frames
$9$ to $12$ relative to its value at the end of the observation, and a hand is released within a chunk when
it is in contact at frame $12$ and out of contact at some frame of the chunk.

\paragraph{The force changes far more often than the contact state.} Within the first predicted chunk the
contact state of a hand changes in \ForceOneFlip{} of cases, and that of the two hands taken together in
only \ForceOneMerged{}. The force of a hand in contact, in contrast, changes by more than $20\%$ in
\ForceOneChangeTwenty{} of cases and by more than $50\%$ in \ForceOneChangeFifty{}, and the center of
pressure moves by more than two cells in \ForceOneCopMove{} (\Cref{tab:force}). Over the second chunk all
of these rates are higher. A binary contact state therefore describes only a small part of how the
interaction evolves over the predicted frames.

\paragraph{The observed course of the force signals whether a grasp will change.} We divide the hands in
contact at frame $12$ into thirds by their observed trend. A hand whose force holds steady is released
within the first predicted chunk in \ForceOneReleaseSteady{} of cases, a hand whose force decays in
\ForceOneReleaseDec{}, and a hand whose force rises in \ForceOneReleaseInc{}. A steady force thus indicates
a grasp that persists, and a changing force, in either direction, a grasp about to change.
This is not a consequence of the force level alone. Within each fifth of the distribution of the current
force, a decaying or rising hand is released \ForceOneRatioMin{} to \ForceOneRatioMax{} times as often as a
steady one. A logistic predictor of release that sees the current force and contact area reaches an area
under the ROC curve of \ForceOneAucBase{}; adding the observed trend raises it to \ForceOneAucFull{}
(\ForceOneAucGain{}, $95\%$ interval \ForceOneAucGainCi{} over episodes, cross-validated with episodes as
groups), and to \ForceTwoAucFull{} from \ForceTwoAucBase{} for the second chunk.

\paragraph{Relation to the tactile inputs of \Cref{tab:input}.} The per-taxel contact input keeps where
each hand is pressed but removes the force and therefore its trend, and the contact bit per hand removes
both. The statistics above show that the removed information is predictive of how the interaction
continues, which is consistent with the order of cost in \Cref{tab:input}. They describe the corpus and
do not by themselves show which part of the reading the model uses.

\begin{table}[h]
\centering
\caption{\textbf{The force and its location change far more often than the contact state, and the observed
force trend signals an imminent release.} Training clips in the evaluation layout; the second and third
blocks use the hands in contact at the last observed frame.}
\label{tab:force}
\footnotesize
\setlength{\tabcolsep}{5pt}
\begin{tabular}{lcc}
\toprule
& First chunk & Second chunk \\
& (frames 13 to 28) & (frames 29 to 44) \\
\midrule
\multicolumn{3}{l}{\emph{Share of predicted chunks in which, relative to the last observed frame, \dots}} \\
\quad the contact state of the two hands taken together changes & 5.0\% & 6.2\% \\
\quad the contact state of a hand changes & 16.8\% & 21.0\% \\
\quad the force of a hand in contact changes by more than $20\%$ & 46.1\% & 60.6\% \\
\quad the force of a hand in contact changes by more than $50\%$ & 22.4\% & 36.4\% \\
\quad the center of pressure of a hand moves by more than two cells & 47.8\% & 62.3\% \\
\midrule
\multicolumn{3}{l}{\emph{Release within the chunk of a hand in contact at the last observed frame, by observed force trend}} \\
\quad decaying (lowest third) & 16.3\% & 16.8\% \\
\quad steady (middle third) & 6.6\% & 9.8\% \\
\quad rising (highest third) & 11.6\% & 17.1\% \\
\midrule
\multicolumn{3}{l}{\emph{Predicting release, area under the ROC curve}} \\
\quad current force and contact area & 0.723 & 0.627 \\
\quad adding the observed force trend & 0.735 & 0.649 \\
\bottomrule
\end{tabular}

\end{table}

\end{document}